# Generating Natural Language Descriptions from OWL Ontologies: the NaturalOWL System

**Ion Androutsopoulos**                                   ION@AUEB.GR
*Department of Informatics,*
*Athens University of Economics and Business, Greece*
*Digital Curation Unit – Institute for the Management of Information Systems,*
*Research Centre "Athena", Athens, Greece*

**Gerasimos Lampouras**                              LAMPOURAS06@AUEB.GR
*Department of Informatics,*
*Athens University of Economics and Business, Greece*

**Dimitrios Galanis**                                  GALANISD@AUEB.GR
*Department of Informatics,*
*Athens University of Economics and Business, Greece*
*Institute for Language and Speech Processing,*
*Research Centre "Athena", Athens, Greece*

## Abstract

We present NaturalOWL, a natural language generation system that produces texts describing individuals or classes of OWL ontologies. Unlike simpler OWL verbalizers, which typically express a single axiom at a time in controlled, often not entirely fluent natural language primarily for the benefit of domain experts, we aim to generate fluent and coherent multi-sentence texts for end-users. With a system like NaturalOWL, one can publish information in OWL on the Web, along with automatically produced corresponding texts in multiple languages, making the information accessible not only to computer programs and domain experts, but also end-users. We discuss the processing stages of NaturalOWL, the optional domain-dependent linguistic resources that the system can use at each stage, and why they are useful. We also present trials showing that when the domain-dependent linguistic resources are available, NaturalOWL produces significantly better texts compared to a simpler verbalizer, and that the resources can be created with relatively light effort.

## 1. Introduction

Ontologies play a central role in the Semantic Web (Berners-Lee, Hendler, & Lassila, 2001; Shadbolt, Berners-Lee, & Hall, 2006). Each ontology provides a conceptualization of a knowledge domain (e.g., consumer electronics) by defining the classes and subclasses of the individuals (entities) in the domain, the types of possible relations between them etc. The current standard to specify Semantic Web ontologies is OWL (Horrocks, Patel-Schneider, & van Harmelen, 2003), a formal language based on description logics (Baader, Calvanese, McGuinness, Nardi, & Patel-Schneider, 2002), RDF, and RDF SCHEMA (Antoniou & van Harmelen, 2008), with OWL2 being the latest version of OWL (Grau, Horrocks, Motik, Par-sia, Patel-Schneider, & Sattler, 2008). Given an OWL ontology for a knowledge domain, one can publish on the Web machine-readable data pertaining to that domain (e.g., catalogues of products, their features etc.), with the data having formally defined semantics based on the conceptualization of the ontology.[1] Following common practice in Semantic Web research, we actually use the term *ontology* to refer jointly to *terminological knowledge* (TBox) that establishes a conceptualization of a knowledge domain, and *assertional knowledge* (ABox) that describes particular individuals.

Several equivalent OWL syntaxes have been developed, but people unfamiliar with formal knowledge representation often have difficulties understanding them (Rector, Drummond, Horridge, Rogers, Knublauch, Stevens, Wang, & Wroe, 2004). For example, the following statement defines the class of St. Emilion wines, using the functional-style syntax of OWL, one of the easiest to understand, which we also adopt throughout this article.[2]

```
EquivalentClasses(:StEmilion
  ObjectIntersectionOf(:Bordeaux
    ObjectHasValue(:locatedIn :stEmilionRegion) ObjectHasValue(:hasColor :red)
    ObjectHasValue(:hasFlavor :strong)          ObjectHasValue(:madeFrom :cabernetSauvignonGrape)
    ObjectMaxCardinality(1 :madeFrom)))
```

To make ontologies easier to understand, several *ontology verbalizers* have been developed (Schwitter, 2010a). Verbalizers usually translate the axioms (in our case, OWL statements) of the ontology one by one to controlled, often not entirely fluent English statements, typically without considering the coherence of the resulting texts, and mostly for the benefit of domain experts. By contrast, in this article we present a system that aims to produce fluent and coherent multi-sentence texts describing classes or individuals of OWL ontologies, with the texts intended to be read by end-users (e.g., customers of on-line retail sites). For example, our system can generate the following text from the OWL statement above, if the ontology has been annotated with domain-dependent linguistic resources discussed below.

St. Emilion is a kind of Bordeaux from the St. Emilion region. It has red color and strong flavor. It is made from exactly one grape variety: Cabernet Sauvignon grapes.

Our system, called NaturalOWL, is open-source and supports both English and Greek. Hence, Greek texts can also be generated from the same OWL statements, as in the following product description, provided that appropriate Greek linguistic resources are also available. By contrast, OWL verbalizers typically produce only English (or English-like) sentences.

```
ClassAssertion(:Laptop :tecraA8)
ObjectPropertyAssertion(:manufacturedBy :tecraA8 :toshiba)
ObjectPropertyAssertion(:hasProcessor :tecraA8 :intelCore2)
DataPropertyAssertion(:hasMemoryInGB :tecraA8 "2"^^xsd:nonNegativeInteger)
DataPropertyAssertion(:hasHardDiskInGB :tecraA8 "110"^^xsd:nonNegativeInteger)
DataPropertyAssertion(:hasSpeedInGHz :tecraA8 "2"^^xsd:float)
DataPropertyAssertion(:hasPriceInEuro :tecraA8 "850"^^xsd:nonNegativeInteger)
```

[English description:] Tecra A8 is a laptop, manufactured by Toshiba. It has an Intel Core 2 processor, 2 GB RAM and a 110 GB hard disk. Its speed is 2 GHz and it costs 850 Euro.

---

1. See `http://owl.cs.manchester.ac.uk/repository/` for a repository of OWL ontologies.
2. Consult `http://www.w3.org/TR/owl2-primer/` for an introduction to the functional-style syntax of OWL.

[Greek description:] Ο Tecra A8 είναι ένας φορητός υπολογιστής, κατασκευασμένος από την Toshiba. Διαθέτει επεξεργαστή Intel Core 2, 2 gb ram και σκληρό δίσκο 110 gb. Η ταχύτητά του είναι 2 ghz και κοστίζει 850 Ευρώ.

The examples above illustrate how a system like NaturalOWL can help publish information on the Web both as OWL statements and as texts generated from the OWL statements. This way, information becomes easily accessible to both computers, which can process the OWL statements, and end-users speaking different languages; and changes in the OWL statements can be automatically reflected in the texts by regenerating them. To produce fluent, coherent multi-sentence texts, NaturalOWL relies on natural language generation (NLG) methods (McKeown, 1985; Reiter & Dale, 2000) to a larger extent compared to existing OWL verbalizers; for example, it includes mechanisms to avoid repeating information, to order the facts to be expressed, aggregate smaller sentences into longer ones, generate referring expressions etc. Although NLG is an established area, this is the first article to discuss in detail an NLG system for OWL ontologies, excluding simpler verbalizers. We do not propose novel algorithms from a theoretical NLG perspective, but we show that there are several particular issues that need to be considered when generating from OWL ontologies. For example, some OWL statements lead to overly complicated sentences, unless they are converted to simpler intermediate representations first; there are also several OWL-specific opportunities to aggregate sentences (e.g., when expressing axioms about the cardinalities of properties); and referring expression generation can exploit the class hierarchy.

NaturalOWL can be used with any OWL ontology, but to obtain texts of high quality *domain-dependent generation resources* are required; for example, the classes of the ontology can be mapped to natural language names, the properties to sentence plans etc. Similar linguistic resources are used in most NLG systems, though different systems adopt different linguistic theories and algorithms, requiring different resources. There is little consensus on exactly what information NLG resources should capture, apart from abstract specifications (Mellish, 2010). The domain-dependent generation resources of NaturalOWL are created by a *domain author*, a person familiar with OWL, when the system is configured for a new ontology. The domain author uses the Protégé ontology editor and a Protégé plug-in that allows editing the domain-dependent generation resources and invoking NaturalOWL to view the resulting texts.[3] We do not discuss the plug-in in this article, since it is very similar to the authoring tool of M-PIRO (Androutsopoulos, Oberlander, & Karkaletsis, 2007).

OWL ontologies often use English words or concatenations of words (e.g., `manufacturedBy`) as identifiers of classes, properties, and individuals. Hence, some of the domain-dependent generation resources can often be extracted from the ontology by guessing, for example, that a class identifier like `Laptop` in our earlier example is a noun that can be used to refer to that class, or that a statement of the form `ObjectPropertyAssertion(:manufacturedBy` $X$ $Y$) should be expressed in English as a sentence of the form "$X$ was manufactured by $Y$". Most OWL verbalizers follow this strategy. Similarly, if domain-dependent generation resources are not provided, NaturalOWL attempts to extract them from the ontology, or it uses generic resources. The resulting texts, however, are of lower quality; also, non-English texts cannot be generated, if the identifiers of the ontology are English-like. There is a tradeoff between reducing the effort to construct domain-dependent generation resources for OWL ontologies, and obtaining higher-quality texts in multiple languages, but this tradeoff has not been investigated in previous work. We present trials we performed to measure the effort required to construct the domain-dependent generation resources of NaturalOWL and the extent to which they improve the resulting texts, also comparing against a simpler verbalizer that requires no domain-dependent generation resources. The trials show that the domain-dependent generation resources help NaturalOWL produce significantly better texts, and that the resources can be constructed with relatively light effort, compared to the effort typically needed to construct an ontology.

---

3. Consult `http://protege.stanford.edu/` for information on Protégé. NaturalOWL and its Protégé plug-in are freely available from `http://nlp.cs.aueb.gr/software.html`. We describe NaturalOWL version 2 in this article; version 1 (Galanis & Androutsopoulos, 2007) used a less principled representation of its domain-dependent generation resources, without supporting OWL2.

Overall, the main contributions of this article are: (i) it is the first detailed discussion of a complete, general-purpose NLG system for OWL ontologies and the particular issues that arise when generating from OWL ontologies; (ii) it shows that a system that relies on NLG methods to a larger extent, compared to simpler OWL verbalizers, can produce significantly better natural language descriptions of classes and individuals, provided that appropriate domain-dependent generation resources are available; (iii) it shows how the descriptions can be generated in more than one languages, again provided that appropriate resources are available; (iv) it shows that the domain-dependent generation resources can be constructed with relatively light effort. As already noted, this article does not present novel algorithms from a theoretical NLG perspective. In fact, some of the algorithms that NaturalOWL uses are of a narrower scope, compared to more fully-fledged NLG algorithms. Nevertheless, the trials show that the system produces texts of reasonable quality, especially when domain-dependent generation resources are provided. We hope that if NaturalOWL contributes towards a wider adoption of NLG methods on the Semantic Web, other researchers may wish to contribute improved components, given that NaturalOWL is open-source.

NaturalOWL is based on ideas from ILEX (O'Donnell, Mellish, Oberlander, & Knott, 2001) and M-PIRO (Isard, Oberlander, Androutsopoulos, & Matheson, 2003). The ILEX project developed an NLG system that was demonstrated mostly with museum exhibits, but did not support OWL.[4] The M-PIRO project produced a multilingual extension of the system of ILEX, which was tested in several domains (Androutsopoulos et al., 2007). Attempts to use the generator of M-PIRO with OWL, however, ran into problems (Androutsopoulos, Kallonis, & Karkaletsis, 2005). By contrast, NaturalOWL was especially developed for OWL.

In the remainder of this article, we assume that the reader is familiar with RDF, RDF SCHEMA, and OWL. Readers unfamiliar with the Semantic Web may wish to consult an introductory text first (Antoniou & van Harmelen, 2008).[5] We also note that the recently very popular Linked Data are published and interconnected using Semantic Web technologies.[6] Most Linked Data currently use only RDF and RDF SCHEMA, but OWL is in effect a superset of RDF SCHEMA and, hence, the work of this paper also applies to Linked Data.

---

4. Dale et al. (1998) and Dannels (2008, 2012) also discuss NLG for museums.

5. A longer version of this article, with more background for readers who are unfamiliar with OWL and the Semantic Web, is available as a technical report (Androutsopoulos, Lampouras, & Galanis, 2012); see `http://nlp.cs.aueb.gr/publications.html`.

6. Consult `http://linkeddata.org/`. See also the work of Duma and Klein (2013).

Section 2 below briefly discusses some related work; we provide further pointers to related work in the subsequent sections. Section 3 then explains how NaturalOWL generates texts, also discussing the domain-dependent generation resources of each processing stage. Section 4 describes the trials we performed to measure the effort required to construct the domain-dependent generation resources and their impact on the quality of the generated texts. Section 5 concludes and proposes future work.

## 2. Related Work

We use the functional-style syntax of OWL in this article, but several equivalent OWL syntaxes exist. There has also been work to develop controlled natural languages (CNLs), mostly English-like, to be used as alternative OWL syntaxes. Sydney OWL Syntax (SOS) (Cregan, Schwitter, & Meyer, 2007) is an English-like CNL with a bidirectional mapping to and from the functional-style syntax of OWL; SOS is based on PENG (Schwitter & Tilbrook, 2004). A similar bidirectional mapping has been defined for Attempto Controlled English (ACE) (Kaljurand, 2007). Rabbit (Denaux, Dimitrova, Cohn, Dolbear, & Hart, 2010) and CLOnE (Funk, Tablan, Bontcheva, Cunningham, Davis, & Handschuh, 2007) are other OWL CNLs, mostly intended to be used by domain experts when authoring ontologies (Denaux, Dolbear, Hart, Dimitrova, & Cohn, 2011). We also note that some OWL CNLs cannot express all the kinds of OWL statements (Schwitter, Kaljurand, Cregan, Dolbear, & Hart, 2008).

Much work on OWL CNLs focuses on ontology authoring and querying (Bernardi, Calvanese, & Thorne, 2007; Kaufmann & Bernstein, 2010; Schwitter, 2010b); the emphasis is mostly on the direction from CNL to OWL or query languages.[7] More relevant to our work are CNLs like SOS and ACE, to which automatic mappings from normative OWL syntaxes are available. By feeding an OWL ontology expressed, for example, in functional-style syntax to a mapping that translates to an English-like CNL, all the axioms of the ontology can be turned into English-like sentences. Systems of this kind are often called *ontology verbalizers*. This term, however, also includes systems that translate from OWL to English-like statements that do not belong in an explicitly defined CNL (Halaschek-Wiener, Golbeck, Parsia, Kolovski, & Hendler, 2008; Schutte, 2009; Power & Third, 2010; Power, 2010; Stevens, Malone, Williams, Power, & Third, 2011; Liang, Stevens, Scott, & Rector, 2011b).

Although verbalizers can be viewed as performing a kind of light NLG, they typically translate axioms one by one, as already noted, without considering the coherence (or topical cohesion) of the resulting texts, usually without aggregating sentences nor generating referring expressions, and often by producing sentences that are not entirely fluent or natural. For example, ACE and SOS occasionally use variables instead of referring expressions (Schwitter et al., 2008). Also, verbalizers typically do not employ domain-dependent generation resources and typically do not support multiple languages. Expressing the exact meaning of the axioms of the ontology in an unambiguous manner is considered more important in verbalizers than composing a fluent and coherent text in multiple languages, partly because the verbalizers are typically intended to be used by domain experts.

---

7. *Conceptual authoring* or WYSIWYM (Power & Scott, 1998; Hallett, Scott, & Power, 2007), which has been applied to OWL (Power, 2009), and *round-trip authoring* (Davis, Iqbal, Funk, Tablan, Bontcheva, Cunningham, & Handschuh, 2008) are bidirectional, but focus mostly on ontology authoring and querying.

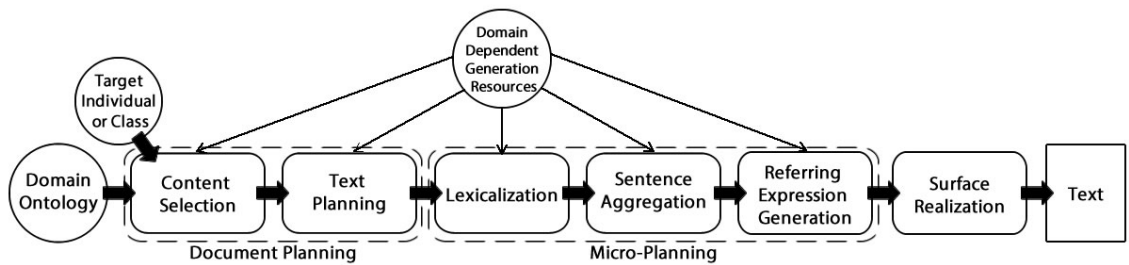

Figure 1: The processing stages and sub-stages of NaturalOWL.

Some verbalizers use ideas and methods from NLG. For example, some verbalizers include sentence aggregation (Williams & Power, 2010) and text planning (Liang, Scott, Stevens, & Rector, 2011a). Overall, however, NLG methods have been used only to a very limited extent with OWL ontologies. A notable exception is ONTOSUM (Bontcheva, 2005), which generates natural language descriptions of individuals, but apparently not classes, from RDF SCHEMA and OWL ontologies. It is an extension of MIAKT (Bontcheva & Wilks, 2004), which was used to generate medical reports. Both were implemented in GATE (Bontcheva, Tablan, Maynard, & Cunningham, 2004) and they provide graphical user interfaces to manipulate domain-dependent generation resources (Bontcheva & Cunningham, 2003). No detailed description of ONTOSUM appears to have been published, however, and the system does not seem to be publicly available, unlike NaturalOWL. Also, no trials of ONTOSUM with independently created ontologies seem to have been published. More information on how ONTOSUM compares to NaturalOWL can be found elsewhere (Androutsopoulos et al., 2012).

Mellish and Sun (2006) focus on lexicalization and sentence aggregation, aiming to produce a single aggregated sentence from an input collection of RDF triples; by contrast, NaturalOWL produces multi-sentence texts. In complementary work, Mellish et al. (2008) consider content selection for texts describing OWL classes. Unlike NaturalOWL, their system does not express only facts that are explicit in the ontology, but also facts deduced from the ontology. Nguyen et al. (2012) discuss how the proof trees of facts deduced from OWL ontologies can be explained in natural language. It would be particularly interesting to examine how deduction and explanation mechanisms could be added to NaturalOWL.

## 3. The Processing Stages and Resources of NaturalOWL

NaturalOWL adopts a pipeline architecture, which is common in NLG (Reiter & Dale, 2000), though the number and purpose of its components often vary (Mellish, Scott, Cahill, Paiva, Evans, & Reape, 2006). Our system generates texts in three stages, *document planning*, *micro-planning*, and *surface realization*, discussed in the following sections; see Figure 1.

### 3.1 Document Planning

Document planning consists of *content selection*, where the system selects the information to convey, and *text planning*, where it plans the structure of the text to be generated.

#### 3.1.1 CONTENT SELECTION

In content selection, the system first retrieves from the ontology all the OWL statements that are relevant to the class or individual to be described, it then converts the selected owl statements to message triples, which are easier to express as sentences, and it finally selects among the message triples the ones to be expressed.

*OWL statements for individual targets*

Let us first consider content selection when NaturalOWL is asked to describe an *individual* (an entity), and let us call that individual the *target*. The system scans the OWL statements of the ontology, looking for statements of the forms listed in the left column of Table 1.[8] In effect, it retrieves all the statements that describe the target directly, as opposed to statements describing another individual or a (named) class the target is related to.

OWL allows arbitrarily many nested `ObjectUnionOf` and `ObjectIntersectionOf` operators, which may lead to statements that are very difficult to express in natural language. To simplify text generation and to ensure that the resulting texts are easy to comprehend, we do not allow nested `ObjectIntersectionOf` and `ObjectUnionOf` operators in the ontologies the texts are generated from. In Table 1, this restriction is enforced by requiring class identifiers to appear at some points where OWL also allows expressions that construct unnamed classes using operators. If an ontology uses unnamed classes at points where Table 1 requires class identifiers (named classes), it can be easily modified to comply with Table 1 by defining new named classes for nested unnamed ones.[9] In practice, nested `ObjectUnionOf` and `ObjectIntersectionOf` operators are rare; see the work of Power et al. (Power, 2010; Power & Third, 2010; Power, 2012) for information of the frequencies of different types of OWL statements.[10]

Statements of the form `ClassAssertion(Class target)` may be quite complex, because `Class` is not necessarily a class identifier. It may also be an expression constructing an unnamed class, as in the following example. This is why there are multiple rows for `ClassAssertion` in Table 1.

```
ClassAssertion(
  ObjectIntersectionOf(:Wine
    ObjectHasValue(:locatedIn :stEmilionRegion)
    ObjectHasValue(:hasColor :red)                    ObjectHasValue(:hasFlavor :strong)
    ObjectHasValue(:madeFrom :cabernetSauvignonGrape) ObjectMaxCardinality(1 :madeFrom))
  :chateauTeyssier2007)
```

NaturalOWL would express the OWL statement above by generating a text like the following.

The 2007 Chateau Teyssier is a wine from the St. Emilion region. It has red color and strong flavor. It is made from exactly one grape variety: Cabernet Sauvignon grapes.

Recall that the texts of NaturalOWL are intended to be read by end-users. Hence, we prefer to generate texts that may not emphasize enough some of the subtleties of the OWL statements, in order to produce more readable texts. An OWL expert might prefer, for example, the following description of `chateauTeyssier2007`, which mirrors more closely the corresponding OWL statements.

---

8. Some OWL statements shown in Table 1 with two arguments can actually have more arguments, but they can be converted to the forms shown.

9. It is also easy to automatically detect nested unnamed classes and replace them, again automatically, by new named classes (classes with OWL identifiers). The domain author would have to be consulted, though, to provide meaningful OWL identifiers for the new classes (otherwise arbitrary identifiers would have to be used) and natural language names for the new classes (see Section 3.2.1 below).

10. One could also refactor some nested operators; for example, $t \in ((A \cup B) \cap (C \cup D))$ is equivalent to $t \in (A \cup B)$ and $t \in (C \cup D)$. The conversion to message triples, to be discussed below, in effect also performs some refactoring, but it cannot cope with all the possible nested union and intersection operators, which is why we disallow them as a general rule.

| OWL statements | Message triples |
|---|---|
| `ClassAssertion(NamedClass target)` | `<target, instanceOf, NamedClass>` |
| `ClassAssertion(`<br>`  ObjectComplementOf(NamedClass) target)` | `<target, not(instanceOf), NamedClass>` |
| `ClassAssertion(`<br>`  ObjectOneOf(indiv1 indiv2 ...) target)` | `<target, oneOf,`<br>`  or(indiv1, indiv2, ...)>` |
| `ClassAssertion(`<br>`  ObjectHasValue(objProp indiv) target)` | `<target, objProp, indiv>` |
| `ClassAssertion(`<br>`  ObjectHasValue(dataProp dataValue) target)` | `<target, dataProp, dataValue>` |
| `ClassAssertion(ObjectHasSelf(objProp) target)` | `<target, objProp, target>` |
| `ClassAssertion(`<br>`  ObjectMaxCardinality(number prop [NamedClass])`<br>`  target)` | `<target, maxCardinality(prop),`<br>`  number[:NamedClass]>` |
| `ClassAssertion(`<br>`  ObjectMinCardinality(number prop [NamedClass])`<br>`  target)` | `<target, minCardinality(prop),`<br>`  number[:NamedClass]>` |
| `ClassAssertion(`<br>`  ObjectExactCardinality(number prop [NamedClass])`<br>`  target)` | `<target, exactCardinality(prop),`<br>`  number[:NamedClass]>` |
| `ClassAssertion(`<br>`  ObjectSomeValuesFrom(objProp NamedClass) target)` | `<target, someValuesFrom(objProp),`<br>`  NamedClass>` |
| `ClassAssertion(`<br>`  ObjectAllValuesFrom(objProp NamedClass) target)` | `<target, allValuesFrom(objProp),`<br>`  NamedClass>` |
| `ClassAssertion(`<br>`  ObjectIntersectionOf(C1 C2 ...) target)` | *convert*`(ClassAssertion(C1 target))`<br>*convert*`(ClassAssertion(C2 target))` ... |
| `ClassAssertion(`<br>`  ObjectUnionOf(C1 C2 ...) target)` | `or(`*convert*`(ClassAssertion(C1 target)),`<br>`    `*convert*`(ClassAssertion(C2 target)),`<br>`    ...)` |
| `ObjectPropertyAssertion(objProp target indiv)` | `<target, objProp, indiv>` |
| `DataPropertyAssertion(dataProp target dataValue)` | `<target, dataProp, dataValue>` |
| `NegativeObjectPropertyAssertion(`<br>`  objProp target indiv)` | `<target, not(objProp), indiv>` |
| `NegativeDataPropertyAssertion(`<br>`  dataProp target dataValue)` | `<target, not(dataProp), dataValue>` |
| `DifferentIndividuals(target indiv)` | `<target, differentIndividuals, indiv>` |
| `DifferentIndividuals(indiv target)` | `<target, differentIndividuals, indiv>` |
| `SameIndividual(target indiv)` | `<target, sameIndividual, indiv>` |
| `SameIndividual(indiv target)` | `<target, sameIndividual, indiv>` |
| Notation: Square brackets indicate optional arguments, and *convert*$(\xi)$ a recursive application of the conversion to $\xi$. `NamedClass` is a class identifier; `objProp`, `dataProp`, and `prop` are identifiers of object properties, datatype properties, and properties; `indiv`, `indiv1`, ... are identifiers of individuals; `dataValue` is a datatype value; and `C`, `C1`, ... are class identifiers, or expressions constructing classes without `ObjectIntersectionOf` or `ObjectUnionOf`. | |

Table 1: OWL statements for an **individual** target, and the corresponding message triples.

The 2007 Chateau Teyssier is a member of the intersection of: (a) the class of wines, (b) the class of individuals from (not necessarily exclusively) the St. Emilion region, (c) the class of individuals that have (not necessarily exclusively) red color, (d) the class of individuals that have (not necessarily exclusively) strong flavor, (e) the class of individuals that are made exclusively from Cabernet Sauvignon grapes.

Stricter texts of this kind, however, seem inappropriate for end-users. In fact, it could be argued that even mentioning that the wine is made from *exactly* one grape variety in the text that NaturalOWL produces is inappropriate for end-users. Our system can be instructed to avoid mentioning this information via user modeling annotations, discussed below.

*OWL statements for class targets*

If the system is asked to describe a *class*, rather than an individual, it scans the ontology for statements of the forms listed in the left column of Table 2. The class to be described must be a named one, meaning that it must have an OWL identifier, and `Target` denotes its identifier. Again, to simplify the generation process and to avoid producing complicated texts, Table 2 requires class identifiers to appear at some points where OWL also allows expressions that construct unnamed classes using operators. If an ontology uses unnamed classes at points where Table 2 requires class identifiers, it can be easily modified.

In texts describing classes, it is difficult to express informally the difference between `EquivalentClasses` and `SubClassOf`. `EquivalentClasses(C1 C2)` means that any individual of `C1` also belongs in `C2`, and vice versa. By contrast, `SubClassOf(C1 C2)` means that any member of `C1` also belongs in `C2`, but the reverse is not necessarily true. If we replace `EquivalentClasses` by `SubClassOf` in the definition of `StEmilion` of page 672, any member of `StEmilion` is still necessarily also a member of the intersection, but a wine with all the characteristics of the intersection is not necessarily a member of `StEmilion`. Consequently, one should perhaps add sentences like the ones shown in italics below, when expressing `EquivalentClasses` and `SubClassOf`, respectively.

St. Emilion is a kind of Bordeaux from the St. Emilion region. It has red color and strong flavor. It is made from exactly one grape variety: Cabernet Sauvignon grapes. *Every St. Emilion has these properties, and anything that has these properties is a St. Emilion.*

St. Emilion is a kind of Bordeaux from the St. Emilion region. It has red color and strong flavor. It is made from exactly one grape variety: Cabernet Sauvignon grapes. *Every St. Emilion has these properties, but something may have these properties without being a St. Emilion.*

NaturalOWL produces the same texts, without the sentences in italics, for both `SubClassOf` and `EquivalentClasses`, to avoid generating texts that sound too formal. Also, it may not mention some of the information of the ontology about a target class (e.g., that a St. Emilion has strong flavor), when user modeling indicates that this information is already known or that the text should not exceed a particular length. Hence, the generated texts express *necessary*, not *sufficient* conditions for individuals to belong in the target class.

*OWL statements for second-level targets*

In some applications, expressing additional OWL statements that are *indirectly* related to the target may be desirable. Let us assume, for example, that the target is the individual `exhibit24`, and that the following directly relevant statements have been retrieved from the ontology. NaturalOWL would express them by generating a text like the one below.

```
ClassAssertion(:Aryballos :exhibit24)
ObjectPropertyAssertion(:locationFound :exhibit24 :heraionOfDelos)
ObjectPropertyAssertion(:creationPeriod :exhibit24 :archaicPeriod)
ObjectPropertyAssertion(:paintingTechniqueUsed :exhibit24 :blackFigureTechnique)
ObjectPropertyAssertion(:currentMuseum :exhibit24 :delosMuseum)
```

| OWL statements | Message triples |
|---|---|
| `EquivalentClasses(Target C)` | *convert*(SubClassOf(`Target C`)) |
| `EquivalentClasses(C Target)` | *convert*(SubClassOf(`Target C`)) |
| `SubClassOf(Target NamedClass)` | <`Target`, isA, `NamedClass`> |
| `SubClassOf(Target ObjectComplementOf(NamedClass))` | <`Target`, not(isA), `NamedClass`> |
| `SubClassOf(Target`<br>`ObjectOneOf(indiv1 indiv2 ...))` | <`Target`, oneOf,<br>or(`indiv1, indiv2, ...`)> |
| `SubClassOf(Target ObjectHasValue(objProp indiv))` | <`Target`, `objProp`, `indiv`> |
| `SubClassOf(Target`<br>`ObjectHasValue(dataProp dataValue))` | <`Target`, `dataProp`, `dataValue`> |
| `SubClassOf(Target ObjectHasSelf(objProp))` | <`Target`, `objProp`, `Target`> |
| `SubClassOf(Target`<br>`ObjectMaxCardinality(number prop [NamedClass]))` | <`Target`, maxCardinality(`prop`),<br>`number`[:`NamedClass`]> |
| `SubClassOf(Target`<br>`ObjectMinCardinality(number prop [NamedClass]))` | <`Target`, minCardinality(`prop`),<br>`number`[:`NamedClass`]> |
| `SubClassOf(Target`<br>`ObjectExactCardinality(number prop [NamedClass]))` | <`Target`, exactCardinality(`objProp`),<br>`number`[:`NamedClass`]> |
| `SubClassOf(Target`<br>`ObjectSomeValuesFrom(objProp NamedClass))` | <`Target`, someValuesFrom(`objProp`),<br>`NamedClass`> |
| `SubClassOf(Target`<br>`ObjectAllValuesFrom(objProp NamedClass))` | <`Target`, allValuesFrom(`objProp`),<br>`NamedClass`> |
| `SubClassOf(Target`<br>`ObjectIntersectionOf(C1 C2 ...))` | *convert*(SubClassOf(`C1 Target`))<br>*convert*(SubClassOf(`C2 Target`)) ... |
| `SubClassOf(Target`<br>`ObjectUnionOf(C1 C2 ...))` | or(*convert*(SubClassOf(`C1 Target`)),<br>    *convert*(SubClassOf(`C2 Target`)),<br>    ...) |
| `DisjointClasses(Target NamedClass)` | <`Target`, not(isA), `NamedClass`> |
| `DisjointClasses(NamedClass Target)` | <`Target`, not(isA), `NamedClass`> |

Notation: Square brackets indicate optional arguments, and *convert*($\xi$) a recursive application of the conversion to $\xi$. `NamedClass` is a class identifier; `objProp`, `dataProp`, and `prop` are identifiers of object properties, datatype properties, and properties; `indiv`, `indiv1`, ... are identifiers of individuals; `dataValue` is a datatype value; and `C`, `C1`, ... are class identifiers, or expressions constructing classes without `ObjectIntersectionOf` or `ObjectUnionOf`.

Table 2: OWL statements for a **class** target, and the corresponding message triples.

This is an aryballos, found at the Heraion of Delos. It was created during the archaic period and it was decorated with the black-figure technique. It is currently in the Museum of Delos.

The names of classes and individuals can be shown as hyperlinks to indicate that they can be used as subsequent targets. Clicking on a hyperlink would be a request to describe the corresponding class or individual. Alternatively, we may retrieve in advance the OWL statements for the subsequent targets and add them to those of the current target.

More precisely, assuming that the target is an individual, the subsequent targets, called *second-level targets*, are the target's class, provided that it is a named one, and the individuals the target is directly linked to via object properties. NaturalOWL considers second-level targets only when the current target is an individual, because with class targets, second-level targets often lead to complicated texts. To retrieve OWL statements for both the current and the second-level targets (when applicable), or only for the current target, we set the *maximum fact distance* to 2 or 1, respectively. Returning to `exhibit24`, let us assume that the maximum fact distance is 2 and that the following OWL statements for second-level targets have been retrieved.[11]

---

11. Consult `http://www.w3.org/TR/owl-time/` for more principled representations of time in OWL.

```
SubClassOf(:Aryballos :Vase)
SubClassOf(:Aryballos
  ObjectHasValue(:exhibitTypeCannedDescription
    "An aryballos was a small spherical vase with a narrow neck, in which the athletes
    kept the oil they spread their bodies with"^^xsd:string))
DatatypePropertyAssertion(:periodDuration :archaicPeriod "700 BC to 480 BC"^^xsd:string)
DatatypePropertyAssertion(:periodCannedDescription :archaicPeriod
    "The archaic period was when the Greek ancient city-states developed"^^xsd:string)
DataPropertyAssertion(:techniqueCannedDescription :blackFigureTechnique
    "In the black-figure technique, the silhouettes are rendered in black on the pale
    surface of the clay, and details are engraved"^^xsd:string)
```

To express all the retrieved OWL statements, including those for the second-level targets, NaturalOWL would now generate a text like the following, which may be preferable, if this is the first time the user encounters an aryballos and archaic exhibits.

> This is an aryballos, a kind of vase. An aryballos was a small spherical vase with a narrow neck, in which the athletes kept the oil they spread their bodies with. This aryballos was found at the Heraion of Delos and it was created during the archaic period. The archaic period was when the Greek ancient city-states developed and it spans from 700 BC to 480 BC. This aryballos was decorated with the black-figure technique. In the black-figure technique, the silhouettes are rendered in black on the pale surface of the clay, and details are engraved. This aryballos is currently in the Museum of Delos.

We note that in many ontologies it is impractical to represent all the information in logical terms. In our example, it is much easier to store the information that "An aryballos was a small...bodies with" as a string, i.e., as a *canned* sentence, rather than defining classes, properties, and individuals for spreading actions, bodies, etc. and generating the sentence from a logical meaning representation. Canned sentences, however, have to be entered in multiple versions, if several languages or user types need to be supported.

*Converting OWL statements to message triples*

Tables 1 and 2 also show how the retrieved OWL statements can be rewritten as triples of the form $\langle S, P, O \rangle$, where $S$ is the target or a second-level target; $O$ is an individual, datatype value, class, or a set of individuals, datatype values, or classes that $S$ is mapped to; and $P$ specifies the kind of mapping. We call $S$ the *semantic subject* or *owner* of the triple, and $O$ the *semantic object* or *filler*; the triple can also be viewed as a field named $P$, owned by $S$, and filled by $O$. For example, the OWL statements about exhibit24 shown above, including those about the second-level targets, are converted to the following triples.

```
<:exhibit24, instanceOf, :Aryballos>
<:exhibit24, :locationFound, :heraionOfDelos>
<:exhibit24, :creationPeriod, :archaicPeriod>
<:exhibit24, :paintingTechniqueUsed, :blackFigureTechnique>
<:exhibit24, :currentMuseum, :delosMuseum>
<:Aryballos, isA, :Vase>
<:Aryballos, :exhibitTypeCannedDescription, "An aryballos was a... bodies with"^^xsd:string>
<:archaicPeriod, :periodDuration, "700 BC to 480 BC"^^xsd:string>
<:archaicPeriod, :periodCannedDescription, "The archaic period was..."^^xsd:string>
<:blackFigureTechnique, :techniqueCannedDescription, "In the black-figure..."^^xsd:string>
```

More precisely, $P$ can be: (i) a property of the ontology; (ii) one of the keywords isA, instanceOf, oneOf, differentIndividuals, sameIndividuals; or (iii) an expression of the form *modifier*($\rho$), where *modifier* may be not, maxCardinality etc. (see Tables 1 and 2) and $\rho$ is a property of the ontology. We hereafter call *properties* all three types of $P$, though types (ii) and (iii) are strictly not properties in the terminology of OWL. When we need to distinguish between the three types, we use the terms *property of the ontology*, *domain-independent property*, and *modified property*, respectively.

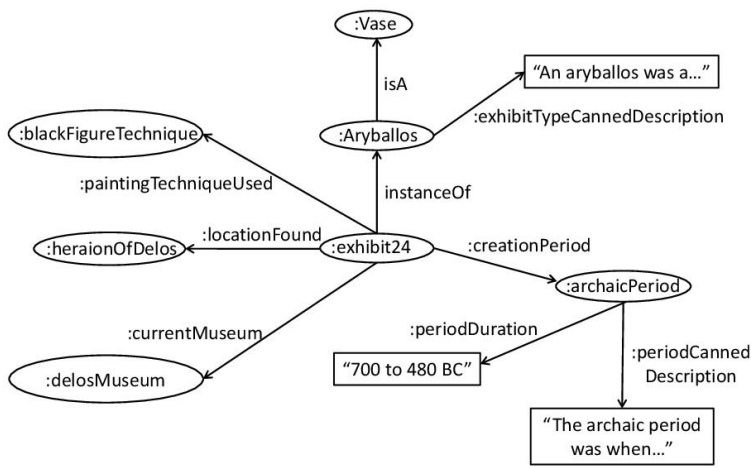

Figure 2: Graph view of message triples.

Every OWL statement or collection of OWL statements can be represented as a set of RDF triples.[12] The triples of Tables 1–2 are similar, but not the same as RDF triples. Most notably, expressions of the form *modifier*($\rho$) cannot be used as $P$ in RDF triples. To avoid confusion, we call *message triples* the triples of Tables 1–2, to distinguish them from RDF triples. As with RDF triples, message triples can be viewed as forming a graph. Figure 2 shows the graph for the message triples of `exhibit24`; the triple linking `blackFigureTechnique` to a canned sentence is not shown to save space. The second-level targets are the classes and individuals at distance one from the target (`exhibit24`).[13] By contrast, the graph for the RDF triples representing the OWL statements would be more complicated, and second-level targets would not always be at distance one from the target.

Each message triple is intended to be easily expressible as a simple sentence, which is not always the case with RDF triples representing OWL statements. The message triples also capture similarities of the sentences to be generated that may be less obvious when looking at the original OWL statements or the RDF triples representing them. For example, the `ClassAssertion` and `SubClassOf` statements below are mapped to identical message triples, apart from the identifiers of the individual and the class, and the similarity of the message triples reflects the similarity of the resulting sentences, also shown below.

```
ClassAssertion(ObjectMaxCardinality(1 :madeFromGrape) :product145)
```

```
<:product145, maxCardinality(:madeFromGrape), 1>
```

Product 145 is made from at most one grape.

---

```
SubClassOf(:StEmilion ObjectMaxCardinality(1 :madeFromGrape))
```

```
<:StEmilion, maxCardinality(:madeFromGrape), 1>
```

St. Emilion is made from at most one grape.

By contrast, without the conversion to message triples, the OWL statements and the RDF triples representing them would lead to more difficult to follow sentences like the following:

Product 145 is a member of the class of individuals that are made from at most one grape.

St. Emilion is a subclass of the class of individuals that are made from at most one grape.

As a further example, Tables 1 and 2 discard `ObjectIntersectionOf` operators, producing multiple message triples instead. For example, the `EquivalentClasses` statement defining `StEmilion` on page 672 would be converted to the following message triples.

```
<:StEmilion, isA, :Bordeaux>
<:StEmilion, :locatedIn, :stEmilionRegion>
<:StEmilion, :hasColor, :red>
<:StEmilion, :hasFlavor, :strong>
<:StEmilion, :madeFromGrape, :cabernetSauvignonGrape>
<:StEmilion, maxCardinality(:madeFromGrape), 1>
```

The resulting message triples correspond to the sentences below, where subsequent references to `StEmilion` have been replaced by pronouns to improve readability; the sentences could also be aggregated into longer ones, as discussed in later sections.

St. Emilion is a kind of Bordeaux. It is from the St. Emilion region. It has red color. It has strong flavor. It is made from Cabernet Sauvignon grape. It is made from at most one grape variety.

By contrast the original OWL statement of page 672 and the RDF triples representing it would lead to the 'stricter' text of page 678, which is inappropriate for end-users, as already noted. Notice, also, that Table 2 converts `EquivalentClasses` and `SubClassOf` statements to identical triples, where $P$ is isA, since NaturalOWL produces the same texts for both kinds of statements, as already discussed.

Tables 1 and 2 also replace `ObjectUnionOf` operators by disjunctions of message triples. The following OWL statement is mapped to the message triple shown below:

```
ClassAssertion(
UnionOf(ObjectHasValue(:hasFlavor :strong) ObjectHasValue(:hasFlavor :medium))
:houseWine)
```

```
or(<:houseWine, :hasFlavor, :strong>, <:houseWine, :hasFlavor, :medium>)
```

which leads to the first sentence below; the sentence can then be shortened during aggregation, leading to the second sentence below.

The house wine has strong flavor or it has medium flavor.

The house wine has strong or medium flavor.

By contrast, the OWL statement and the corresponding RDF triples in effect say that:

> The house wine is a member of the union of: (i) the class of all wines that have strong flavor, and (ii) the class of all wines that have medium flavor.

*Interest scores and repetitions*

Expressing all the message triples of all the retrieved OWL statements is not always appropriate. Let us assume, for example, that the maximum fact distance is 2 and that a description of `exhibit24` of Figure 2 has been requested by a museum visitor. It may be the case that the visitor has already encountered other archaic exhibits, and that the duration of the archaic period was mentioned in previous descriptions. Repeating the duration of the period may, thus, be undesirable. We may also want to exclude message triples that are uninteresting to particular types of users. For example, there may be message triples providing bibliographic references, which children would probably find uninteresting.

NaturalOWL provides mechanisms allowing the domain author to assign an importance score to every possible message triple, and possibly different scores for different user types (e.g., adults, children). The score is a non-negative integer indicating how interesting a user of the corresponding type will presumably find the information of the message triple, if the information has not already been conveyed to the user. In the museum projects NaturalOWL was originally developed for, the interest scores ranged from 0 (completely uninteresting) to 3 (very interesting), but a different range can also be used. The scores can be specified for all the message triples that involve a particular property $P$ (e.g., $P = $ `madeFrom`), or for all the message triples that involve semantic subjects $S$ of a particular class (e.g., $S \in$ `Statue` or $S = $ `Statue`) and a particular property $P$, or for message triples that involve particular semantic subjects (e.g., $S = $`exhibit37`) and a particular property $P$. For example, we may wish to specify that the materials of the exhibits in a collection are generally of medium interest ($P = $ `madeFrom`, score 2), that the materials of statues are of lower interest ($S \in$ `statue`, $P = $ `madeFrom`, score 1), perhaps because all the statues of the collection are made from stone, but that the material of the particular statue `exhibit24` is very important ($S = $ `exhibit10`, $P = $ `madeFrom`, score 3), perhaps because `exhibit24` is a gold statue.

We do not discuss the mechanisms that can be used to assign interest scores to message triples in this article, but a detailed description of these mechanisms can be found elsewhere (Androutsopoulos et al., 2012). We also note that when human-authored texts describing individuals and classes of the ontology are available along with the OWL statements or, more generally, the logical facts they express, statistical and machine learning methods can be employed to learn to automatically select or assign interest scores to logical facts (Duboue & McKeown, 2003; Barzilay & Lapata, 2005; Kelly, Copestake, & Karamanis, 2010). Another possibility (Demir, Carberry, & McCoy, 2010) would be to compute the interest scores with graph algorithms like PageRank (Brin & Page, 1998).

The domain author can also specify how many times each message triple has to be repeated, before it can be assumed that users of different types have assimilated it. Once a triple has been assimilated, it is never repeated in texts for the same user. For example, the domain author can specify that children assimilate the duration of a historical period when it has been mentioned twice; hence, the system may repeat, for example, the duration of the archaic period in two texts. NaturalOWL maintains a personal model for each end-user. The model shows which message triples were conveyed to the particular user in previous texts, and how many times. Again, more information about the user modeling mechanisms of NaturalOWL can be found elsewhere (Androutsopoulos et al., 2012).

*Selecting the message triples to convey*

When asked to describe a target, NaturalOWL first retrieves from the ontology the relevant OWL statements, possibly also for second-level targets. It then converts the retrieved statements to message triples, and consults their interest scores and the personal user models to rank the message triples by decreasing interest score, discarding triples that have already been assimilated. If a message triple about the target has been assimilated, then all the message triples about second-level targets that are connected to the assimilated triple are also discarded; for example, if the `creationPeriod` triple (edge) of Figure 2 has been assimilated, then the triples about the archaic period (the edges leaving from `archaicPeriod`) are also discarded. The system then selects up to `maxMessagesPerPage` triples from the most interesting remaining ones; `maxMessagesPerPage` is a parameter whose value can be set to smaller or larger values for types of users that prefer shorter or longer texts, respectively.

*Limitations of content selection*

OWL allows one to define the broadest possible domain and range of a particular property, using statements like the following.

```
ObjectPropertyDomain(:madeFrom :Wine)  ObjectPropertyRange(:madeFrom :Grape)
```

In practice, more specific range restrictions are then imposed for particular subclasses of the property's domain. For example, the following statements specify that when `madeFrom` is used with individuals from the subclass `GreekWine` of `Wine`, the range (possible values) of `madeFrom` should be restricted to individuals from the subclass `GreekGrape` of `Grape`.

```
SubClassOf(:GreekWine :Wine)  SubClassOf(:GreekGrape :Grape)
SubClassOf(:GreekWine AllValuesFrom(:madeFrom :GreekGrape))
```

NaturalOWL considers `AllValuesFrom` and similar restrictions (see Tables 1 and 2), but not `ObjectPropertyDomain` and `ObjectPropertyRange` statements. The latter typically provide too general and, hence, uninteresting information from the perspective of end-users.

More generally, NaturalOWL does not consider OWL statements that express axioms about properties, meaning statements declaring that a property is symmetric, asymmetric, reflexive, irreflexive, transitive, functional, that its inverse is functional, that a property is the inverse of, or disjoint with another property, that it is subsumed by a chain of other properties, or that it is a subproperty (more specific) of another property. Statements of this kind are mostly useful in consistency checks, in deduction, or when generating texts describing the properties themselves (e.g., what being a grandparent of somebody means).[14]

### 3.1.2 Text Planning

For each target, the previous mechanisms produce the message triples to be expressed, with each triple intended to be easily expressible as a single sentence. The text planner of NaturalOWL then orders the message triples, in effect ordering the corresponding sentences.

---

14. Subproperties without sentence plans, discussed below, could inherit sentence plans from their superproperties, but in that case we automatically extract sentence plans from the ontology instead.

*Global and local coherence*

When considering *global coherence*, text planners attempt to build a structure, usually a tree, that shows how the clauses, sentences, or larger segments of the text are related to each other, often in terms of rhetorical relations (Mann & Thompson, 1998). The allowed or preferred orderings of the sentences (or segments) often follow, at least partially, from the global coherence structure. In the texts, however, that NaturalOWL is intended to generate, the global coherence structures tend to be rather uninteresting, because most of the sentences simply provide additional information about the target or the second-level targets, which is why global coherence is not considered in NaturalOWL.[15]

When considering *local coherence*, text planners usually aim to maximize measures that examine whether or not adjacent sentences (or segments) continue to focus on the same entities or, if the focus changes, how smooth the transition is. Many local coherence measures are based on Centering Theory (CT) (Grosz, Joshi, & Weinstein, 1995; Poesio, Stevenson, & Di Eugenio, 2004). Consult the work of Karamanis et al. (2009) for an introduction to CT and a CT-based analysis M-PIRO's texts, which also applies to the texts of NaturalOWL.

When the maximum fact distance of NaturalOWL is 1, all the sentence-to-sentence transitions are of a type known in CT as CONTINUE, which is the preferred type. If the maximum fact distance is 2, however, the transitions are not always CONTINUE. We repeat below the long aryballos description of page 681 without sentence aggregation. For readers familiar with CT, we show in italics the most salient noun phrase of each sentence $u_n$, which realizes the discourse entity known as the *preferred center* $c_p(u_n)$. The underlined noun phrases realize the *backward looking center* $c_b(u_n)$, roughly speaking the most salient discourse entity of the previous sentence that is also mentioned in the current sentence.

(1) *This* (exhibit) is an aryballos. (2) *An aryballos* is a kind of vase. (3) *An aryballos* was a small spherical vase with a narrow neck, in which the athletes kept the oil they spread their bodies with. ● (4) *This aryballos* was found at the Heraion of Delos. (5) *It* was created during the archaic period. (6) *The archaic period* was when the Greek ancient city-states developed. (7) *It* spans from 700 BC to 480 BC. ● (8) *This aryballos* was decorated with the black-figure technique. (9) In *the black-figure technique*, the silhouettes are rendered in black on the pale surface of the clay, and details are engraved. ● (10) *This aryballos* is currently in the Museum of Delos.

In sentence 4, where $c_p(u_4)$ is the target exhibit, $c_b(u_4)$ is undefined and the transition from sentence 3 to 4 is a NOCB, a type of transition to be avoided; we mark NOCB transitions with bullets. In sentence 6, $c_p(u_6) = c_b(u_6) \neq c_b(u_5)$, and we have a kind of transition known as SMOOTH-SHIFT (Poesio et al., 2004), less preferred than CONTINUE, but better than NOCB. Another NOCB occurs from sentence 7 to 8, followed by a SMOOTH-SHIFT from sentence 8 to 9, and another NOCB from sentence 9 to 10. All the other transitions are CONTINUE.

The text planner of NaturalOWL groups together sentences (message triples) that describe a particular second-level target (e.g., sentences 2–3, 6–7, and 9) and places each group immediately after the sentence that introduces the corresponding second-level target (immediately after sentences 1, 5, and 8). Thus the transition from a sentence that introduces a second-level target to the first sentence that describes the second-level target (e.g., from sentence 1 to 2, from 5 to 6, from 8 to 9) is a smooth-shift (or a continue in the special case from the initial sentence 1 to 2). A nocb occurs only at sentences that return to providing information about the primary target, after a group of sentences that provide information about a second-level target. All the other transitions are of type continue.

---

15. Liang et al. (2011a) and Power (2011) seem to agree that very few rhetorical relations are relevant when generating texts from OWL ontologies.

A simple strategy to avoid nocb transitions would be to end the generated text once all the message triples that describe a second-level target have been reported, and record in the user model that the other message triples that content selection provided were not actually conveyed. In our example, this would generate sentences 1 to 3; then if the user requested more information about the exhibit, sentences 4 to 7 would be generated etc.

*Topical order*

When ordering sentences, we also need to consider the topical similarity of adjacent sentences. Compare, for example, the following two texts.

{$_{locationSection}$ The Stoa of Zeus Eleutherios is located in the western part of the Agora. It is located next to the Temple of Apollo Patroos.} {$_{buildSection}$ It was built around 430 bc. It was built in the Doric style. It was built out of porous stone and marble.} {$_{useSection}$ It was used during the Classical period, the Hellenistic period, and the Roman period. It was used as a religious place and a meeting point.} {$_{conditionSection}$ It was destroyed in the late Roman period. It was excavated in 1891 and 1931. Today it is in good condition.}

The Stoa of Zeus Eleutherios was built in the Doric style. It was excavated in 1891 and 1931. It was built out of porous stone and marble. It is located in the western part of the Agora. It was destroyed in the late Roman period. It was used as a religious place and a meeting point. It is located next to the Temple of Apollo Patroos. It was built around 430 bc. Today it is in good condition. It was used during the Classical period, the Hellenistic period, and the Roman period.

Even though both texts contain the same sentences, the second text is more difficult to follow, if at all acceptable. The first one is better, because it groups together topically related sentences. We mark the sentence groups in the first text by curly brackets, but the brackets would not be shown to end-users. In longer texts, sentence groups may optionally be shown as separate paragraphs or sections, which is why we call them *sections*.

To allow the message triples (and the corresponding sentences) to be grouped by topic, the domain author may define sections (e.g., `locationSection`, `buildSection`) and assign each property to a single section (e.g., assign the properties `isInArea` and `isNextTo` to `locationSection`). Each message triple is then placed in the section of its property. An ordering of the sections and of the properties inside each section can also be specified, causing the message triples to be ordered accordingly (e.g., we may specify that `locationSection` should precede `buildSection`, and that inside `locationSection`, the `isInArea` property should be expressed before `isNextTo`). The sections, the assignments of the properties to sections, and the order of the sections and the properties are defined in the domain-dependent generation resources (Androutsopoulos et al., 2012).

*The overall text planning algorithm*

NaturalOWL's text planning algorithm is summarized in Figure 3. If the message triples to be ordered include triples that describe second-level targets, i.e., triples $\langle S, P, O \rangle$ whose owner $S$ is a second-level target, then the triples of the primary and each second-level target

```
procedure orderMessageTriples
inputs:
  t[0]: primary target
  t[1], ..., t[n]: second-level targets
  L[0]: unordered list of triples describing t[0]
  ...
  L[n]: unordered list of triples describing t[n]
  SMap: mapping from properties to sections
  SOrder: partial order of sections
  POrder: partial order of properties within sections
output:
  ordered list of message triples
steps:
  for i := 0 to n { orderMessageTriplesAux(L[i], SMap, SOrder, POrder) }
  for i := 1 to n { insertAfterFirst(<t[0], _, t[i]>, L[0], L[i]) }
  return L[0]

procedure orderMessageTriplesAux
inputs:
  L: unordered list of triples about a single target
  SMap: mapping from properties to sections
  SOrder: partial order of sections
  POrder: partial order of properties within sections
local variables:
  S[1], ..., S[k]: lists, each with triples of one section
output:
  ordered list of message triples about a single target
steps:
  <S[1], ..., S[k]> := splitInSections(L, SMap)
  for i := 1 to k { S[i] := orderTriplesInSection(S[i], POrder) }
  <S[1], ..., S[k]> := reorderSections(S[1], ..., S[k], SOrder)
  return concatenate(S[1], ..., S[k])
```

Figure 3: The overall text planning algorithm of NaturalOWL.

are ordered separately, using the ordering of properties and sections. The ordered triples of each second-level target are then inserted into the ordered list of the primary target triples, immediately after the first triple that introduces the second-level target, i.e., immediately after the first triple whose $O$ is the second-level target.

*Further related work on text planning*

The ordering of properties and sections is similar to *text schemata* (McKeown, 1985), roughly speaking domain-dependent patterns that specify the possible arrangements of different types of sentences (or segments). Sentence ordering has been studied extensively in text summarization (Barzilay, Elhadad, & McKeown, 2002). Duboue and McKeown (2001) discuss methods that could be used to learn the order of sentences or other segments in NLG from semantically tagged training corpora. Consult also the work of Barzilay and Lee (2004), Elsner et al. (2007), Barzilay and Lapata (2008), and Chen et al. (2009).

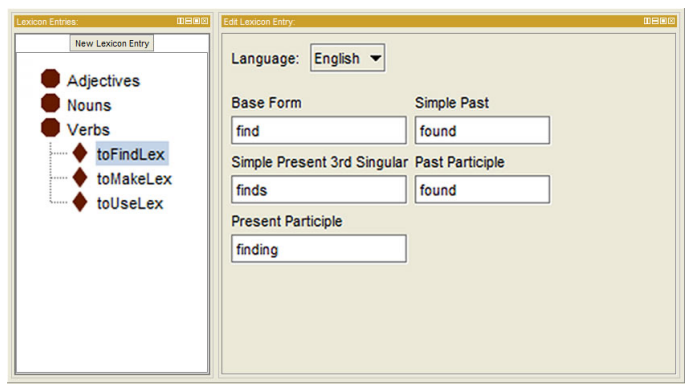

Figure 4: A lexicon entry for the verb "to find".

## 3.2 Micro-planning

The processing stages we have discussed so far select and order the message triples to be expressed. The next stage, *micro-planning*, consists of three sub-stages: *lexicalization*, *sentence aggregation*, and *generation of referring expressions*; see also Figure 1 on page 676.

### 3.2.1 LEXICALIZATION

During lexicalization, NLG systems usually turn the output of content selection (in our case, the message triples) to abstract sentence specifications. In NaturalOWL, for every property of the ontology and every supported natural language, the domain author may specify one or more template-like sentence plans to indicate how message triples involving that property can be expressed. We discuss below how sentence plans are specified, but first a slight deviation is necessary, to briefly discuss the lexicon entries of NaturalOWL.

*Lexicon entries*

For each verb, noun, or adjective that the domain author wishes to use in the sentence plans, a lexicon entry has to be provided, to specify the inflectional forms of that word.[16] All the lexicon entries are multilingual (currently bilingual); this allows sentence plans to be reused across similar languages when no better option is available, as discussed elsewhere (Androutsopoulos et al., 2007). Figure 4 shows the lexicon entry for the verb whose English base form is "find", as viewed by the domain author when using the Protégé plug-in of NaturalOWL. The identifier of the lexicon entry is `toFindLex`. The English part of the entry shows that the base form is "find", the simple past is "found" etc. Similarly, the Greek part of the lexicon entry would show the base form of the corresponding verb ("βρίσκω") and its inflectional forms in the various tenses, persons etc. The lexicon entries for nouns and adjectives are very similar.

Most of the English inflectional forms could be automatically produced from the base forms by using simple morphology rules. We hope to exloit an existing English morphology component, such as that of SIMPLENLG (Gatt & Reiter, 2009), in future work. Similar morphology rules for Greek were used in the authoring tool of M-PIRO (Androutsopoulos et al., 2007), and we hope to include them in a future version of NaturalOWL. Rules of this kind would reduce the time a domain author spends creating lexicon entries. In the ontologies we have considered, however, a few dozens of lexicon entries for verbs, nouns, and adjectives suffice. Hence, even without facilities to automatically produce inflectional forms, creating the lexicon entries is rather trivial. Another possibility would be to exploit a general-purpose lexicon or lexical database, like WordNet (Fellbaum, 1998) or CELEX, though resources of this kind often do not cover the highly technical concepts of ontologies.[17]

---

16. No lexicon entries need to be provided for closed-class words, like determiners and prepositions.

The lexicon entries and, more generally, all the domain-dependent generation resources of NaturalOWL are stored as instances of an OWL ontology (other than the ontology the texts are generated from) that describes the linguistic resources of the system (Androutsopoulos et al., 2012). The domain author, however, interacts with the plug-in and does not need to be aware of the OWL representation of the resources. By representing the domain-dependent generation resources in OWL, it becomes easier to publish them on the Web, check them for inconsistencies etc., as with other OWL ontologies.

*Sentence plans*

In NaturalOWL, a sentence plan is a sequence of slots, along with instructions specifying how to fill them in. Figure 5 shows an English sentence plan for the property `usedDuringPeriod`, as viewed by the domain author when using the Protégé plug-in of NaturalOWL. The sentence plan expresses message triples of the form $\langle S, \texttt{usedDuringPeriod}, O \rangle$ by producing sentences like the following.

[$_{slot_1}$This stoa] [$_{slot_2}$was used] [$_{slot_3}$during] [$_{slot_4}$the Classical period].

[$_{slot_1}$The Stoa of Zeus Eleutherios] [$_{slot_2}$was used] [$_{slot_3}$during] [$_{slot_4}$the Classical period, the Hellenistic period, and the Roman period].

The first slot of the sentence plan of Figure 5 is to be filled in with an automatically generated referring expression for the owner $(S)$ of the triple. For example, if the triple to express is `<:stoaZeusEleutherios, :usedDuringPeriod, :classicalPeriod>`, an appropriate referring expression for $S$ may be a demonstrative noun phrase like "this stoa", a pronoun ("it"), or the monument's natural language name ("the Stoa of Zeus Eleutherios"). We discuss the generation of referring expressions below, along with mechanisms to specify natural language names. The sentence plan also specifies that the referring expression must be in nominative case (e.g., "it" or "this stoa", as opposed to the genitive case expressions "its" or "this stoa's", as in "This stoa's height is 5 meters").

The second slot is to be filled in with a form of the verb whose lexicon identifier is `toUseVerb`. The verb form must be in the simple past and passive voice, in positive polarity (as opposed to "was *not* used"). Its number must agree with the number of the expression in the first slot; for example, we want to generate "The Stoa of Zheus Eleutherios *was* used", but "Stoas *were* used". The third slot is filled in with the preposition "during". The fourth slot is filled in with an expression for the filler $(O)$ of the message triple, in accusative case.[18] With `<:stoaZeusEleutherios, :usedDuringPeriod, :classicalPeriod>`, the slot would be filled in with the natural language name of `classicalPeriod`.[19] The sentence plan also allows the resulting sentence to be aggregated with other sentences.

---

17. See `http://www.ldc.upenn.edu/Catalog/catalogEntry.jsp?catalogId=LDC96L14` for CELEX.

18. English prepositions usually require noun phrase complements in accusative (e.g., "on him"). In Greek and other languages, cases have more noticeable effects.

19. Future versions of NaturalOWL may allow a referring expression for $O$ other than its natural language name to be produced (e.g., a pronoun), as with $S$.

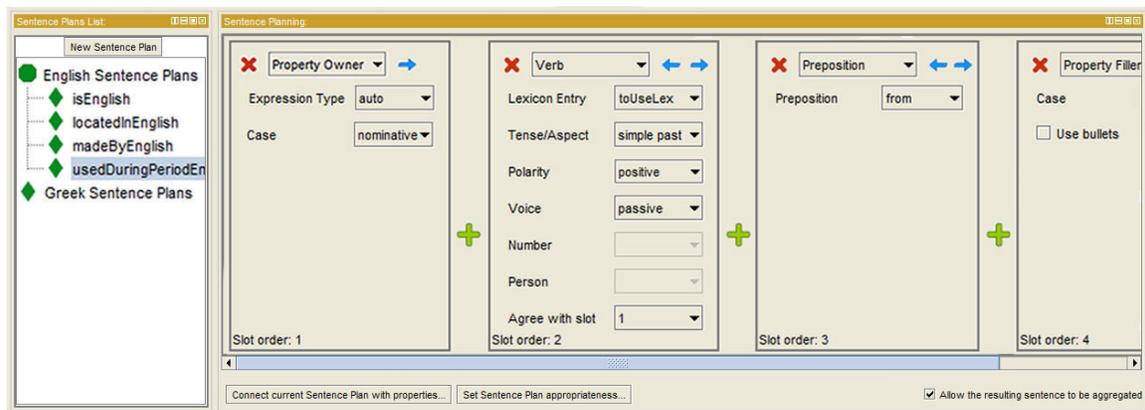

Figure 5: A sentence plan for the property `usedDuringPeriod`.

More generally, the instructions of a sentence plan may indicate that a slot should be filled in with one of the following (i–vii):

(i) *A referring expression for the S* (owner) of the message triple. A sentence plan may specify a particular type of referring expression to use (e.g., always use the natural language name of *S*) or, as in the example of Figure 5, it may allow the system to automatically produce the most appropriate type of referring expression depending on the context.

(ii) *A verb* for which there is a lexicon entry, in a particular form, possibly a form that agrees with another slot. The polarity of the verb can also be manually specified or, if the filler (*O*) of the message triple is a Boolean value, the polarity can be automatically set to match that value (e.g., to produce "It does *not* have a built-in flash" when *O* is `false`).

(iii) *A noun or adjective* from the lexicon, in a particular form (e.g., case, number), or in a form that agrees with another slot.

(iv) *A preposition* or (v) *a fixed string.*

(vi) *An expression for the O* (filler) of triple. If *O* is an individual or class, then the expression is the natural language name of *O*; if *O* is a datatype value (e.g., an integer), then the value itself is inserted in the slot; and similarly if *O* is a disjunction or conjunction of datatype values, individuals, or classes.

(vii) *A concatenation of property values of O*, provided that *O* is an individual. For example, we may need to express a message triple like the first one below, whose (anonymous in the RDF sense) object `_:n` is linked to both a numeric value (via `hasAmount`) and an individual standing for the currency (via `hasCurrency`).

```
<:tecra8, :hasPrice, _:n>   <_:n, :hasAmount, "850"^^xsd:float>   <_:n, :hasCurrency, :euroCurrency>
```

We would want the sentence plan to include a slot filled in with the concatenation of the `hasAmount` value of `_:n` and the natural language name of the `hasCurrency` value of `_:n` (e.g., "850 Euro" in English, "850 Ευρώ" in Greek).

*Default sentence plan*

If no sentence plan has been provided for a particular property of the ontology, NaturalOWL uses a default sentence plan, consisting of three slots. The first slot is filled in with an automatically generated referring expression for the owner (*S*) of the triple, in nominative case. The second slot is filled in with a tokenized form of the OWL identifier of the property. The third slot is filled in with an appropriate expression for the filler (*O*) of the triple, as discussed above, in accusative case (if applicable). For the following message triple, the default sentence plan would produce the sentence shown below:

```
<:stoaZeusEleutherios, :usedDuringPeriod, and(:classicalPeriod, :hellenisticPeriod, :romanPeriod)>
```

Stoa zeus eleutherios used during period classical period, hellenistic period, and roman period.

Notice that we use a single message triple with an `and(...)` filler, instead of a different triple for each period. This kind of triple merging is in effect a form of aggregation, discussed below, but it takes place during content selection. Also, we assumed in the sentence above that the natural language names of the individuals have not been provided either; in this case, NaturalOWL uses tokenized forms of the OWL identifiers of the individuals instead. The tokenizer of NaturalOWL can handle both CamelCase (e.g., `:usedDuringPeriod`) and underscore style (e.g., `:used_during_period`). When other styles are used in the identifiers of properties, classes, and individuals, the output of the tokenizer may be worse than the example suggests, but the resulting sentences can be improved by providing sentence plans and by associating classes and individuals with natural language names, discussed below.

*Using rdfs:label strings*

OWL properties (and other elements of OWL ontologies) can be labeled with strings in multiple natural languages using the `rdfs:label` annotation property, defined in the RDF and OWL standards. For example, the `usedDuringPeriod` property could be labeled with "was used during" as shown below; there could be similar labels for Greek and other languages.

```
AnnotationAssertion(rdfs:label :usedDuringPeriod "was used during"@en)
```

If an `rdfs:label` string has been specified for the property of a message triple, NaturalOWL uses that string in the second slot of the default sentence plan. The quality of the resulting sentences can, thus, be improved, if the `rdfs:label` strings are more natural phrases than the tokenized property identifiers. With the `rdfs:label` shown above, the default sentence plan would produce the following sentence.

Stoa zeus eleutherios was used during classical period, hellenistic period, and roman period.

Even with `rdfs:label` strings, the default sentence plan may produce sentences with disfluencies. Also, the `rdfs:label` strings do not indicate the grammatical categories of their words, and this does not allow the system to apply many of the sentence aggregation rules discussed below. A further limitation of the default sentence plan is that it does not allow the slots for $S$ and $O$ to be preceded or followed, respectively, by any other phrase.

*Sentence plans for domain-independent and modified properties*

The domain author does not need to provide sentence plans for domain-independent properties (e.g., `instanceOf`, `isA`, see Tables 1–2). These properties have fixed, domain-independent semantics; hence, built-in sentence plans are used. The English built-in sentence plans, which also serve as further examples of sentence plans, are summarized in Table 3; the Greek built-in sentence plans are similar. To save space we show the sentence plans as templates in Table 3, and we do not show the sentence plans for negated domain-independent properties (e.g., `not(isA)`), which are similar. Additional slot restrictions not shown in Figure 3 require, for example, subject-verb number agreement and the verb forms ("is" or "was") to be in present tense. Information provided when specifying the natural language names of individuals and classes, discussed below, shows if definite or indefinite articles or no articles at all should be used (e.g., "*the* N97 mini", "exhibit 24", "*a* St. Emilion" or "*the* St. Emilion" or simply "St. Emilion"), and what the default number of each name is (e.g., "A wine color is" or "Wine colors are"). It is also possible to modify the built-in sentence plans; for example, in a museum context we may wish to generate "An aryballos *was* a kind of vase" instead of "An aryballos *is* a kind of vase".

| Forms of message triples and the corresponding built-in sentence plans | Example message triples and possible resulting sentences |
|---|---|
| <$S$, `instanceOf`, $O$><br>$ref(S)$ `toBeVerb` $name(indef, O)$ | `<:eos450d, instanceOf, :PhotographicCamera>`<br>The EOS 450D is a photographic camera. |
| <$S$, `instanceOf`, $O$><br>$ref(S)$ `toBeVerb` $name(adj, O)$ | `<:eos450d, instanceOf, :Cheap>`<br>The EOS 450D is cheap. |
| <$S$, `oneOf`, $O$><br>$ref(S)$ `toBeVerb` $name(O)$ | `<:WineColor, oneOf, or(:white, :rose, :red)>`<br>A wine color is white, rose, or red. |
| <$S$, `differentIndividuals`, $O$><br>$ref(S)$ `toBeVerb` not identical to $name(O)$ | `<:n97, differentIndividuals, :n97mini>`<br>The N97 is not identical to the N97 mini. |
| <$S$, `sameIndividual`, $O$><br>$ref(S)$ `toBeVerb` identical to $name(O)$ | `<:eos450d, sameIndividual, :rebelXSi>`<br>It is identical to the Rebel XSI. |
| <$S$, `isA`, $O$><br>$ref(S)$ `toBeVerb` a kind of $name(noarticle, O)$ | `<:StEmilion, isa, :Bordeaux>`<br>St. Emilion is a kind of Bordeaux. |
| <$S$, `isA`, $O$><br>$ref(S)$ `toBeVerb` $name(adj, O)$ | `<:StEmilion, isa, :Red>`<br>St. Emilion is red. |
| Notation: $ref(\xi)$ stands for a referring expression for $\xi$; $name(\xi)$ is the natural language name of $\xi$; $name(indef, \xi)$ and $name(noarticle, \xi)$ mean that the name should be a noun phrase with an indefinite or no article. Sentence plans involving $name(adj, \xi)$ are used when the natural language name of $\xi$ is a sequence of one or more adjectives; otherwise the sentence plan of the previous row is used. | |

Table 3: Built-in English sentence plans for **domain-independent properties**.

The sentence plans for modified properties (e.g., `minCardinality(manufacturedBy)`, see Tables 1–2) are also automatically produced, from the sentence plans of the unmodified properties (e.g., `manufacturedBy`).

*Specifying the appropriateness of sentence plans*

Multiple sentence plans may be provided for the same property of the ontology and the same language. Different appropriateness scores (similar to the interest scores of properties) can then be assigned to alternative sentence plans per user type. This allows specifying, for example, that a sentence plan that generates sentences like "This amphora depicts Milti-ades" is less appropriate when interacting with children, compared to an alternative sentence plan with a more common verb (e.g., "shows"). Automatically constructed sentence plans inherit the appropriateness scores of the sentence plans they are constructed from.

*Related work on sentence plans*

The sentence plans of NaturalOWL are similar to expressions of sentence planning languages like SPL (Kasper & Whitney, 1989) that are used in generic surface realizers, such as FUF/SURGE (Elhadad & Robin, 1996), KPML (Bateman, 1997), REALPRO (Lavoie & Ram-bow, 1997), NITROGEN/HALOGEN (Langkilde, 2000), and OPENCCG (White, 2006). The sentence plans of NaturalOWL, however, leave fewer decisions to subsequent stages. This has the disadvantage that our sentence plans often include information that could be obtained from large-scale grammars or corpora (Wan, Dras, Dale, & Paris, 2010). On the other hand, the input to generic surface realizers often refers to non-elementary linguistic concepts (e.g., features of a particular syntax theory) and concepts of an *upper model* (Bateman, 1990); the latter is a high-level domain-independent ontology that may use a very different conceptualization than the ontology the texts are to be generated from. Hence, linguistic expertise, for example in Systemic Grammars (Halliday, 1994) in the case of KPML (Bateman, 1997), and effort to understand the upper model are required. By contrast, the sentence plans of NaturalOWL require the domain author to be familiar with only elementary linguistic concepts (e.g., tense, number), and they do not require familiarity with an upper model. Our sentence plans are simpler than, for example, the templates of Busemann and Horacek (1999) or McRoy et al. (2003), in that they do not allow, for instance, conditionals or recursive invocation of other templates. See also the work of Reiter (1995) and van Deemter et al. (2005) for a discussion of template-based vs. more principled NLG.

When corpora of texts annotated with the message triples they express are available, templates can also be automatically extracted (Ratnaparkhi, 2000; Angeli, Liang, & Klein, 2010; Duma & Klein, 2013). Statistical methods that jointly perform content selection, lexicalization, and surface realization have also been proposed (Liang, Jordan, & Klein, 2009; Konstas & Lapata, 2012a, 2012b), but they are currently limited to generating single sentences from flat records.

*Specifying natural language names*

The domain author can assign *natural language* (NL) names to the individuals and named classes of the ontology; recall that by named classes we mean classes that have OWL identifiers. If an individual or named class is not assigned an NL name, then its `rdfs:label` or a tokenized form of its identifier is used instead. The NL names that the domain author provides are specified much as sentence plans, i.e., as sequences of slots. For example, we may specify that the English NL name of the class `ItalianWinePiemonte` is the concatenation of the following slots; we explain the slots below.

[$_{indef}$ an] [$_{adj}$ Italian] [$_{headnoun}$ wine] [$_{prep}$ from] [$_{def}$ the] [$_{noun}$ Piemonte] [$_{noun}$ region]

This would allow NaturalOWL to generate the sentence shown below from the following message triple; a tokenized form of the identifier of `wine32` is used.

```
<:wine32, instanceOf, :ItalianWinePiemonte>
```

Wine 32 is an Italian wine from the Piemonte region.

Similarly, we may assign the following NL names to the individuals `classicalPeriod`, `stoa`, `ZeusEleutherios`, `gl2011`, and the classes `ComputerScreen` and `Red`. NaturalOWL makes no distinction between common and proper nouns; both are entered as nouns in the lexicon, and may be multi-word (e.g., "Zeus Eleutherios"). NaturalOWL can also be instructed to capitalize the words of particular slots (e.g., "Classical").

[*def* the] [*adj* Classical] [*headnoun* period] , [*def* the] [*headnoun* stoa] [*prep* of] [*noun* Zeus Eleutherios],

[*headnoun* GL-2011] , [*indef* a] [*noun* computer] [*headnoun* screen] , [*headadj* red]

These NL names could be used to express the message triples shown below:

```
<:stoaZeusEleutherios, :usedDuringPeriod,  :classicalPeriod>
```

The Stoa of Zeus Eleutherios was used during the Classical period.

```
<:gl2011, instanceOf, :ComputerScreen>  <:gl2011, instanceOf, :Red>
```

GL-2011 is a computer screen. GL-2011 is red.

More precisely, each NL name is a sequence of slots, with accompanying instructions specifying how the slots are to be filled in. Each slot can be filled in with:

(i) *An article*, definite or indefinite. The article in the first slot (if present) is treated as the article of the overall NL name.

(ii) *A noun or adjective flagged as the head* (main word) of the NL name. Exactly one head must be specified per NL name and it must have a lexicon entry. The number and case of the head, which is also taken to be the number and case of the overall NL name, can be automatically adjusted per context. For example, different sentence plans may require the same NL name to be in nominative case when used as a subject, but in accusative when used as the object of a verb; and some aggregation rules, discussed below, may require a singular NL name to be turned into plural. Using the lexicon entries, which list the inflectional forms of nouns and adjectives, NaturalOWL can adjust the NL names accordingly. The gender of head adjectives can also be automatically adjusted, whereas the gender of head nouns is fixed and specified by their lexicon entries.

(iii) *Any other noun or adjective*, among those listed in the lexicon. The NL name may require a particular inflectional form to be used, or it may require an inflectional form that agrees with another slot of the NL name.

(iv) *A preposition*, or (v) *any fixed string*.

As with sentence plans, the domain author specifies NL names by using the Protégé plug-in of NaturalOWL. Multiple NL names can be specified for the same individual or class, and they can be assigned different appropriateness scores per user type; hence, different terminology (e.g., common names of diseases) can be used when generating texts for non-experts, as opposed to texts for experts (e.g., doctors). The domain author can also specify, again using the plug-in, if the NL names of particular individuals or classes should involve definite, indefinite, or no articles, and if the NL names should be in singular or plural by default. For example, we may prefer the texts to mention the class of aryballoi as a single particular generic object, or by using an indefinite singular or plural form, as shown below.

The aryballos is a kind of vase. An aryballos is a kind of vase. Aryballoi are a kind of vase.

### 3.2.2 Sentence Aggregation

The sentence plans of the previous section lead to a separate sentence for each message triple. NLG systems often aggregate sentences into longer ones to improve readability. In NaturalOWL, the maximum number of sentences that can be aggregated to form a single longer sentence is specified per user type via a parameter called `maxMessagesPerSentence`. In the museum contexts our system was originally developed for, setting `maxMessagesPerSentence` to 3 or 4 led to reasonable texts for adult visitors, whereas a value of 2 was used for children. The sentence aggregation of NaturalOWL is performed by a set of manually crafted rules, intended to be domain-independent. We do not claim that this set of rules, which was initially based on the aggregation rules of M-PIRO (Melengoglou, 2002), is complete, and we hope it will be extended in future work; see, for example, the work of Dalianis (1999) for a rich set of aggregation rules.[20] Nevertheless, the current rules of NaturalOWL already illustrate several aggregation opportunities that arise when generating texts from OWL ontologies.

To save space, we discuss only English sentence aggregation; Greek aggregation is similar. We show mostly example *sentences* before and after aggregation, but the rules actually operate on *sentence plans* and they also consider the message triples being expressed. The rules are intended to aggregate short single-clause sentences. Sentence plans that produce more complicated sentences may be flagged (using the tickbox at the bottom of Figure 5) to signal that aggregation should not affect their sentences. The aggregation rules apply almost exclusively to sentences that are adjacent in the ordering produced by the text planner; the only exception are aggregation rules that involve sentences about cardinality restrictions. Hence, depending on the ordering of the text planner there may be more or fewer aggregation opportunities; see the work of Cheng and Mellish (2000) for related discussion. Also, the aggregation rules of NaturalOWL operate on sentences of the same topical section, because aggregating topically unrelated sentences often sounds unnatural.

The aggregation of NaturalOWL is greedy. For each of the rules discussed below, starting from those discussed first, the system scans the original (ordered) sentences from first to last, applying the rule wherever possible, provided that the rule's application does not lead to a sentence expressing more than `maxMessagesPerSentence` original sentences. If a rule can be applied in multiple ways, for example to aggregate two or three sentences, the application that aggregates the most sentences without violating `maxMessagesPerSentence` is preferred.

*Avoid repeating a noun with multiple adjectives:* Message triples of the form $\langle S, P, O_1 \rangle, \ldots,$ $\langle S, P, O_n \rangle$ will have been aggregated into a single message triple $\langle S, P, \text{and}(O_1, \ldots, O_n) \rangle$. If the NL names of $O_1, \ldots, O_n$ are, apart from possible initial determiners, sequences of adjectives followed by the same head noun, then the head noun does not need to be repeated. Let us consider the following message triple. Assuming that the NL names of the three periods are as in the first sentence below, the original sentence will repeat "period" three times. The aggregation rule omits all but the last occurrence of the head noun.

```
<:stoaZeusEleutherios, :usedDuringPeriod, and(:classicalPeriod, :hellenisticPeriod, :romanPeriod)>
```

It was used during the Classical period, the Hellenistic period, and the Roman period. ⇒ It was used during the Classical, the Hellenistic, and the Roman period.

---

20. When appropriate corpora are available, it may also be possible to train aggregation modules (Walker, Rambow, & Rogati, 2001; Barzilay & Lapata, 2006).

*Cardinality restrictions and values:* This is a set of rules that aggregate *all* the sentences (not necessarily adjacent) that express message triples of the form $\langle S, M(P), O \rangle$ and $\langle S, P, O \rangle$, for the same $S$ and $P$, with $M$ being any of `minCardinality`, `maxCardinality`, `exactCardinality`. When these rules are applied, `MaxMessagesPerSentence` is ignored. For example, these rules perform aggregations like the following.

> Model 35 is sold in at most three countries. Model 35 is sold in at least three countries. Model 35 is sold in Spain, Italy, and Greece. $\Rightarrow$ Model 35 is sold in exactly three countries: Spain, Italy, and Greece.

*Class and passive sentence:* This rule aggregates (i) a sentence expressing a message triple $\langle S, \texttt{instanceOf}, C \rangle$ or $\langle S, \texttt{isA}, C \rangle$ and (ii) a passive immediately subsequent sentence expressing a single triple of the form $\langle S, P, O \rangle$, for the same $S$, where $P$ is an (unmodified) property of the ontology. The subject and auxiliary verb of the second sentence are omitted.

> Bancroft Chardonnay is a kind of Chardonnay. It is made in Bancroft. $\Rightarrow$ Bancroft Chardonnay is a kind of Chardonnay made in Bancroft.

*Class and prepositional phrase:* The second sentence now involves the verb "to be" in the active simple present, immediately followed by a preposition; the other conditions are as in the previous rule. The subject and verb of the second sentence are omitted.

> Bancroft Chardonnay is a kind of Chardonnay. It is from Bancroft. $\Rightarrow$ Bancroft Chardonnay is a kind of Chardonnay from Bancroft.

*Class and multiple adjectives:* This rule aggregates (i) a sentence of the same form as in the previous two rules, and (ii) one or more immediately preceding or subsequent sentences, each expressing a single message triple $\langle S, P_i, O_i \rangle$, for the same $S$, where $P_i$ are (unmodified) properties of the ontology. Each of the preceding or subsequent sentences must involve the verb "to be" in the active simple present, immediately followed by only an adjective. The adjectives are absorbed into sentence (i) maintaining their order.

> This is a motorbike. It is red. It is expensive. $\Rightarrow$ This is a red, expensive motorbike.

*Same verb conjunction/disjunction:* In a sequence of sentences involving the same verb form, each expressing a single message triple $\langle S, P_i, O_i \rangle$, where $S$ is the same in all the triples and $P_i$ are (unmodified) properties of the ontology, a conjunction can be formed by mentioning the subject and verb once. The "and" is omitted when a preposition follows.

> It has medium body. It has moderate flavor. $\Rightarrow$ It has medium body and moderate flavor.
> He was born in Athens. He was born in 1918. $\Rightarrow$ He was born in Athens in 1918.

A similar rule applies to sentences produced from disjunctions of message triples, as illustrated below. A variant of the first aggregation rule is then also applied.

> The house wine has strong flavor or it has medium flavor. $\Rightarrow$ The house wine has strong flavor or medium flavor. $\Rightarrow$ The house wine has strong or medium flavor.

*Different verbs conjunction:* When there is a sequence of sentences, not involving the same verb form, each expressing a message triple $\langle S, P_i, O_i \rangle$, where $S$ is the same in all the triples and $P_i$ are (unmodified) properties of the ontology, a conjunction can be formed:

Bancroft Chardonnay is dry. It has moderate flavor. It comes from Napa. $\Rightarrow$ Bancroft Chardonnay is dry, it has moderate flavor, and it comes from Napa.

### 3.2.3 GENERATING REFERRING EXPRESSIONS

A sentence plan may require a referring expression to be generated for the $S$ of a message triple $\langle S, P, O \rangle$. Depending on the context, it may be better, for example, to use the NL name of $S$ (e.g., "the Stoa of Zeus Eleutherios"), a pronoun (e.g., "it"), a demonstrative noun phrase (e.g., "this stoa") etc. Similar alternatives could be made available for $O$, but NaturalOWL currently uses $O$ itself, if it is a datatype value; or the NL name of $O$, its tokenized identifier, or its `rdfs:label`, if $O$ is an entity or class; and similarly for conjunctions and disjunctions in $O$. Hence, below we focus only on referring expressions for $S$.

NaturalOWL currently uses a limited range of referring expressions, which includes only NL names (or tokenized identifiers or `rdfs:label` strings), pronouns, and noun phrases involving only a demonstrative and the NL name of a class (e.g., "this vase"). For example, referring expressions that mention properties of $S$ (e.g., "the vase from Rome") are not generated. Although the current referring expression generation mechanisms of NaturalOWL work reasonably well, they are best viewed as placeholders for more elaborate algorithms (Krahmer & van Deemter, 2012), especially algorithms based on description logics (Areces, Koller, & Striegnitz, 2008; Ren, van Deemter, & Pan, 2010).

Let us consider the following generated text, which expresses the triples $\langle S_i, P_i, O_i \rangle$ shown below. We do not aggregate sentences in this section, to illustrate more cases where referring expressions are needed; aggregation would reduce, however, the number of pronouns, making the text less repetitive. For readers familiar with CT (Section 3.1.2), we show again in italics the noun phrase realizing $c_p(u_n)$, we show underlined the noun phrase realizing $c_b(u_n)$, and we mark NOCB transitions with bullets.

(1) *Exhibit 7* is a statue. (2) <u>It</u> was sculpted by Nikolaou. (3) <u>*Nikolaou*</u> was born in Athens. (4) <u>He</u> was born in 1918. (5) <u>He</u> died in 1998. • (6) *Exhibit 7* is now in the National Gallery. (7) <u>It</u> is in excellent condition.

```
<:exhibit7, instanceOf, :Statue>          <:exhibit7, :hasSculptor, :nikolaou>
<:nikolaou, :cityBorn, :athens>           <:nikolaou, :yearBorn, "1918"^^xsd:integer>
<:nikolaou, :yearDied, "1998"^^xsd:integer> <:exhibit7, :currentLocation, :nationalGallery>
<:exhibit7, :currentCondition, :excellentCondition>
```

NaturalOWL pronominalizes $S_n$ (for $n > 1$) only if $S_n = S_{n-1}$, as in sentences 2, 4, 5, and 7. Since typically $c_p(u_i) = S_i$, we obtain $c_p(u_n) = c_p(u_{n-1})$, whenever $S_n$ is pronominalized, if the pronoun is resolved by the reader as intended. People tend to prefer readings where $c_p(u_n) = c_p(u_{n-1})$, if no other restriction is violated (e.g., gender, number, world knowledge). This helps the pronouns that NaturalOWL generates to be correctly resolved by readers, even when they would appear to be potentially ambiguous. For example, the pronoun of sentence 7 is most naturally understood as referring to the exhibit, as it is intended to, not the gallery, even though both are neuter and can be in excellent condition.

Note that with both referents, the transition from sentence 6 to 7 is a continue; hence, transition type preferences play no role. The gender of each generated pronoun is the gender of the (most appropriate) nl name of the $S$ that the pronoun realizes.[21] If $S$ does not have an nl name, NaturalOWL uses the gender of the (most appropriate) nl name of the most specific class that includes $S$ and has an nl name (or one of these classes, if they are many). nl names can also be associated with *sets* of genders, which give rise to pseudo-pronouns like "he/she"; this may be desirable in the nl name of a class like `Person`.

With some individuals or classes, we may not wish to use nl names, nor tokenized identifiers or `rdfs:label` strings. This is common, for example, in museum ontologies, where some exhibits are known by particular names, but many other exhibits are anonymous and their owl identifiers are not particularly meaningful. NaturalOWL allows the domain author to mark individuals and classes as *anonymous*, to indicate that their nl names, tokenized identifiers, and `rdfs:label` strings should be avoided. When the primary target is marked as anonymous, NaturalOWL uses a demonstrative noun phrase (e.g., "this statue") to refer to it. The demonstrative phrase involves the nl name of the most specific class that subsumes the primary target, has an nl name, and has not been marked as anonymous. Especially in sentences that express `isA` or `instanceOf` message triples about the primary target, the demonstrative phrase is simply "this", to avoid generating sentences like "This statue is a statue". The marking of anonymous individuals and classes currently affects only the referring expressions of the primary target.

### 3.3 Surface Realization

In many nlg systems, the sentences at the end of micro-planning are underspecified; for example, the order of their constituents or the exact forms of their words may be unspecified. Large-scale grammars or statistical models can then be used to fill in the missing information during surface realization, as already discussed (Section 3.2.1). By contrast, in NaturalOWL (and most template-based nlg systems) the (ordered and aggregated) sentence plans at the end of micro-planning already completely specify the surface (final) form of each sentence. Hence, the surface realization of NaturalOWL is mostly a process of converting internal, but fully specified and ordered sentence specifications to the final text. Punctuation and capitalization are also added. Application-specific markup (e.g., html tags, hyperlinks) or images can also be added by modifying the surface realization code of NaturalOWL.

## 4. Trials

In our previous work, NaturalOWL was used mostly to describe cultural heritage objects. In the xenios project, it was tested with an owl version of an ontology that was created during m-piro to document approximately 50 archaeological exhibits (Androutsopoulos et al., 2007).[22] The owl version comprised 76 classes, 343 individuals (including cities, persons etc.), and 41 properties. In xenios, NaturalOWL was also embedded in a robotic avatar that presented the exhibits of m-piro in a virtual museum (Oberlander, Karakatsiotis,

---

21. In languages like Greek that use grammatical instead of natural genders, the pronouns' genders cannot be determined by consulting the ontology (e.g., to check if the referent is animate or inanimate).

22. xenios was co-funded by the European Union and the Greek General Secretariat of Research and Technology; see `http://www.ics.forth.gr/xenios/`.

Isard, & Androutsopoulos, 2008). More recently, in the INDIGO project, NaturalOWL was embedded in mobile robots acting as tour guides in an exhibition about the ancient Agora of Athens.[23] An OWL ontology documenting 43 monuments was used; there were 49 classes, 494 individuals, and 56 properties in total.

In XENIOS and INDIGO, the texts of NaturalOWL were eventually indistinguishable from human-authored texts. We participated, however, in the development of the ontologies, and we may have biased them towards choices (e.g., classes, properties) that made it easier for NaturalOWL to generate high-quality texts. Hence, in the trials discussed below, we wanted to experiment with independently developed ontologies. We also wanted to experiment with different domains, as opposed to cultural heritage.

A further goal was to compare the texts of NaturalOWL against those of a simpler verbalizer. We used the OWL verbalizer of the SWAT project (Stevens et al., 2011; Williams, Third, & Power, 2011), which we found to be particularly robust and useful.[24] The verbalizer produces an alphabetical glossary with an entry for each named class, property, and individual, without requiring domain-dependent generation resources. Each glossary entry is a sequence of English-like sentences expressing the corresponding OWL statements of the ontology. The SWAT verbalizer uses a predetermined partial order of statements in each glossary entry; for example, when describing a class, statements about equivalent classes or super-classes are mentioned first, and individuals belonging in the target class are mentioned last.[25] The verbalizer actually translates the OWL ontology to Prolog, it extracts lexicon entries from OWL identifiers and `rdfs:label` strings, and it uses predetermined sentence plans specified as a DCG grammar. It also aggregates, in effect, message triples of the same property that share one argument ($S$ or $O$) (Williams & Power, 2010).

Our hypothesis was that the domain-dependent generation resources would help NaturalOWL produce texts that end-users would consider more fluent and coherent, compared to those produced by the SWAT verbalizer, but also those produced by NaturalOWL without domain-dependent generation resources. We also wanted to demonstrate that high-quality texts could be produced in both English and Greek, and to measure the effort required to create the domain-dependent generation resources of NaturalOWL for existing ontologies. This effort had not been measured in our previous work, because the development of the domain-dependent generation resources was combined with the development of the ontologies. Since the time needed to create the domain-dependent generation resources depends on one's familiarity with NaturalOWL and its Protégé plug-in, exact times are not particularly informative. Instead, we report figures such as the number of sentence plans, lexicon entries etc. that were required, along with approximate times. We do not evaluate the usability of the Protégé plug-in of NaturalOWL, since it is very similar to the authoring tool of M-PIRO. Previous experiments (Androutsopoulos et al., 2007) showed that computer science graduates with no expertise in NLG could learn to use effectively the authoring tool of M-PIRO to create the necessary domain-dependent generation resources for existing or new ontologies, after receiving the equivalent of a full-day introduction course.

---

23. INDIGO was an FP6 IST project of the European Union; consult `http://www.ics.forth.gr/indigo/`. Videos of the robots of XENIOS and INDIGO are available at `http://nlp.cs.aueb.gr/projects.html`. Two AUEB students, G. Karakatsiotis and V. Pterneas, won the Interoperability Challenge of the 2011 Microsoft Imagine Cup with a similar mobile phone application, called Touring Machine, which uses NaturalOWL; see `http://www.youtube.com/watch?v=PaNAmNC7dZw`.

24. The SWAT verbalizer can be used on-line at `http://swat.open.ac.uk/tools/`. We used the general-purpose version that was on-line in July and August 2011; a similar verbalizer from OWL to ACE (Section 2) is available at `http://attempto.ifi.uzh.ch/site/docs/owl_to_ace.html`. A domain-specific version of SWAT for the SNOMED biomedical ontology has also been developed (Liang et al., 2011a, 2011b).

25. The verbalizer also organizes the English-like sentences of each glossary entry under sub-headings like 'Definition', 'Taxonomy', 'Description', 'Distinctions' (Williams et al., 2011). We discarded these sub-headings, whose meanings were not entirely clear to us, but we retained the order of the sentences.

## 4.1 Trials with the Wine Ontology

In the first trial, we experimented with the Wine Ontology, which is often used in Semantic Web tutorials.[26] It comprises 63 wine classes, 52 wine individuals, a total of 238 classes and individuals (including wineries, regions, etc.), and 14 properties.

We submitted the Wine Ontology to the SWAT verbalizer to obtain its glossary of English-like descriptions of classes, properties, and individuals. We retained only the descriptions of the 63 wine classes and the 52 wine individuals. Subsequently, we also discarded 20 of the 63 wine class descriptions, as they were for trivial classes (e.g., RedWine) and they were stating the obvious (e.g., "A red wine is defined as a wine that has as color Red").[27] In the descriptions of the remaining 43 wine classes and 52 wine individuals, we discarded sentences expressing axioms that NaturalOWL does not consider, for example sentences providing examples of individuals that belong in a class being described. The remaining sentences express the same OWL statements that NaturalOWL expresses when its maximum fact distance is set to 1. Two examples of texts produced by the SWAT verbalizer follow.

Chenin Blanc (class): A chenin blanc is defined as something that is a wine, is made from grape the Chenin Blanc Grape, and is made from grape at most one thing. A chenin blanc both has as flavor Moderate, and has as color White. A chenin blanc both has as sugar only Off Dry and Dry, and has as body only Full and Medium.

The Foxen Chenin Blanc (individual): The Foxen Chenin Blanc is a chenin blanc. The Foxen Chenin Blanc has as body Full. The Foxen Chenin Blanc has as flavor Moderate. The Foxen Chenin Blanc has as maker Foxen. The Foxen Chenin Blanc has as sugar Dry. The Foxen Chenin Blanc is located in the Santa Barbara Region.

Subsequently, we generated texts for the 43 classes and 52 individuals using NaturalOWL without domain-dependent generation resources, hereafter called NaturalOWL(−), setting the maximum fact distance to 1; the resulting texts were very similar to SWAT's.

We then constructed the domain-dependent generation resources of NaturalOWL for the Wine Ontology. The resources are summarized in Table 4. They were constructed by the second author, who devoted three days to their construction, testing, and refinement.[28] Our experience is that it takes weeks (if not longer) to develop an OWL ontology the size of the Wine Ontology (acquire domain knowledge, formulate the axioms in OWL, check for inconsistencies, populate the ontology with individuals etc.); hence, a period of a few days is relatively light effort, compared to the time needed to develop an OWL ontology of this size. Only English texts were generated in this trial; hence, no Greek resources were constructed. We defined only one user type, and we used interest scores only to block sentences stating the obvious, by assigning zero interest scores to the corresponding message triples; we also set `maxMessagesPerSentence` to 3. Only 7 of the 14 properties of the Wine Ontology are used in the OWL statements that describe the 43 classes and 52 individuals. We defined only 5 sentence plans, as some of the 7 properties could be expressed by the same sentence plans. We did not define multiple sentence plans per property. We also assigned the 7 properties to 2 sections, and ordered the sections and properties. We created NL names only when the automatically extracted ones were causing disfluencies. The extracted NL names were obtained from the OWL identifiers of classes and individuals; no `rdfs:label` strings were available. To reduce the number of manually constructed NL names further, we declared the 52 individual wines to be anonymous (and provided no NL names for them). Most of the 67 lexicon entries are used in the remaining 41 NL names of classes and individuals; NL names were very simple, having 2 slots on average. We used NaturalOWL with the domain-dependent resources, hereafter called NaturalOWL(+), to re-generate the 95 texts, again setting the maximum fact distance to 1; example texts follow.

---

26. See http://www.w3.org/TR/owl-guide/wine.rdf.

27. Third (2012) discusses how OWL axioms leading to undesirable sentences of this kind might be detected.

28. Some of the resources were constructed by editing directly their OWL representations, rather than using the Protégé plug-in, which was not fully functional at that time. By using the now fully functional plug-in, the time to create the domain-dependent generation resources would have been shorter.

| Resources | English | Greek |
|---|---|---|
| Sections | 2 | |
| Property assignments to sections | 7 | |
| Interest score assignments | 8 | |
| Sentence plans | 5 | — |
| Lexicon entries | 67 | — |
| Natural language names | 41 | — |

Table 4: Domain-dependent generation resources created for the **Wine Ontology**.

Chenin Blanc (class): A Chenin Blanc is a moderate, white wine. It has only a full or medium body. It is only off-dry or dry. It is made from exactly one wine grape variety: Chenin Blanc grapes.

The Foxen Chenin Blanc (individual): This wine is a moderate, dry Chenin Blanc. It has a full body. It is made by Foxen in the Santa Barbara County.

The resulting 285 texts ($95 \times 3$) of the three systems (SWAT verbalizer, NaturalOWL($-$), NaturalOWL($+$)) were shown to 10 computer science students (both undergraduates and graduate students), who were not involved in the development of NaturalOWL; they were all fluent in English, though not native English speakers, and they did not consider themselves wine experts. The students were told that a glossary of wines was being developed for people who were interested in wines and knew basic wine terms (e.g., wine colors, wine flavors), but who were otherwise not wine experts. Each one of the 285 texts was given to exactly one student. Each student was given approximately 30 texts, approximately 10 randomly selected texts from each system. The OWL statements that the texts were generated from were not shown, and the students did not know which system had generated each text. Each student was shown all of his/her texts in random order, regardless of the system that had generated them. The students were asked to score each text by stating how strongly they agreed or disagreed with statements $S_1$–$S_5$ below. A scale from 1 to 3 was used (1: disagreement, 2: ambivalent, 3: agreement).

| Criteria | SWAT | NaturalOWL($-$) | NaturalOWL($+$) |
|---|---|---|---|
| Sentence fluency | $2.00 \pm 0.15$ | $1.76 \pm 0.15$ | **$2.80 \pm 0.10$** |
| Referring expressions | $1.40 \pm 0.13$ | $1.15 \pm 0.09$ | **$2.72 \pm 0.13$** |
| Text structure | $2.15 \pm 0.16$ | $2.20 \pm 0.16$ | **$2.94 \pm 0.05$** |
| Clarity | $2.66 \pm 0.13$ | $2.55 \pm 0.13$ | **$2.74 \pm 0.11$** |
| Interest | $2.30 \pm 0.15$ | $2.14 \pm 0.16$ | **$2.68 \pm 0.12$** |

Table 5: Results for texts generated from the **Wine Ontology** by the SWAT verbalizer and NaturalOWL with ($+$) and without ($-$) domain-dependent generation resources.

($S_1$) *Sentence fluency*: The sentences of the text are fluent, i.e., each sentence *on its own* is grammatical and sounds natural. When two or more smaller sentences are combined to form a single, longer sentence, the resulting longer sentence is also grammatical and sounds natural.

($S_2$) *Referring expressions*: The use of pronouns and other referring expressions (e.g., "this wine") is appropriate. The choices of referring expressions (e.g., when to use a pronoun or other expression instead of the name of an object) sound natural, and it is easy to understand what these expressions refer to.

($S_3$) *Text structure*: The order of the sentences is appropriate. The text presents information by moving reasonably from one topic to another.

($S_4$) *Clarity*: The text is easy to understand, provided that the reader is familiar with basic wine terms.

($S_5$) *Interest*: People interested in wines, but who are not wine experts, would find the information interesting. Furthermore, there are no redundant sentences in the text (e.g., sentences stating the obvious).[29]

$S_5$ assesses *content selection*, the first processing sub-stage; we expected the differences across the three systems to be very small, as they all reported the same information, with the exception of redundant sentences blocked by using zero interest assignments in NaturalOWL. $S_3$ assesses *text planning*, the second sub-stage; again we expected small differences, as many of the wine properties can be mentioned in any order, though there are some properties (e.g., maker, location) that are most naturally reported separately from others (e.g., color, flavor), which is why we used two sections (Table 4). $S_1$ assesses *lexicalization* and *aggregation*; we decided not to use separate statements for these two stages, since it might have been difficult for the students to understand exactly when aggregation takes place. $S_2$ assesses *referring expression* generation. $S_4$ measures the overall perceived clarity of the texts. There was no statement for *surface realization*, as this stage had a rather trivial effect.

Table 5 shows the average scores of the three systems, with averages computed on the 95 texts of each system, along with 95% confidence intervals (of sample means). For each criterion, the best score is shown in bold; the confidence interval of the best score is also shown in bold if it does not overlap with the other confidence intervals.[30]

As expected, the domain-dependent generation resources clearly help NaturalOWL produce more fluent sentences and much better referring expressions. The text structure scores show that the assignment of the ontology's properties to sections and the ordering of the sections and properties had a greater impact on the perceived structure of the texts than we expected. The highest score of the SWAT verbalizer was obtained in the clarity criterion, which agrees with our experience that one can usually understand what the texts of the SWAT verbalizer mean, even if their sentences are often not entirely fluent, not particularly well ordered, and keep repeating proper names. NaturalOWL($+$)had the highest clarity score, but the difference from the SWAT verbalizer, which had the second highest score, is not statistically significant. NaturalOWL(+)also obtained higher interest scores than the other two systems, with statistically significant differences from both; these differences, which are larger than we expected, can only be attributed to the zero interest score assignments of the domain-dependent generation resources, which blocked sentences stating the obvious, because otherwise all three systems report the same information.

---

29. The students were told not to consider whether or not *additional* information should have been included.

30. When two intervals do not overlap, the difference is statistically significant. When they overlap, the difference may still be statistically significant; we performed paired two-tailed $t$-tests ($\alpha = 0.05$) in these cases. In a pilot study, we also measured the inter-annotator agreement of two of the students on a sample of 30 texts (10 from each system). Agreement was very high (sample Pearson correlation $r \geq 0.91$) in all five criteria. A similar pilot study was performed in the next trial, also indicating very high agreement.

The SWAT verbalizer obtained higher scores than NaturalOWL(−), with the text structure score being the only exception. Only the difference in the referring expression scores of the two systems, though, is statistically significant. Both systems, however, received particularly low scores for their referring expressions, which is not surprising, given that they both always refer to individuals and classes by extracted names; the slightly higher score of the SWAT verbalizer is probably due to its better tokenization of OWL identifiers.

## 4.2 Trials with the Consumer Electronics Ontology

In the second trial, we experimented with the Consumer Electronics Ontology, an OWL ontology for consumer electronics products and services.[31] The ontology comprises 54 classes and 441 individuals (e.g., printer types, paper sizes, manufacturers), but no information about particular products. We added 60 individuals describing 20 digital cameras, 20 camcorders, and 20 printers. The 60 individuals were randomly selected from a publicly available dataset of 286 digital cameras, 613 camcorders, and 58 printers, whose instances comply with the Consumer Electronics Ontology.[32]

We submitted the Consumer Electronics Ontology with the additional 60 individuals to the SWAT verbalizer, and retained only the descriptions of the 60 individuals. Again, we removed sentences expressing axioms NaturalOWL does not consider. We also renamed the string values of some datatype properties to make the texts easier to understand (e.g., "CMT" became "cm"). An example description follows.

The Sony Cyber-shot DSC-T90 is a digital camera.

The Sony Cyber-shot DSC-T90 has as manufacturer Sony.

The Sony Cyber-shot DSC-T90 has as data interface type Usb2 0.

The Sony Cyber-shot DSC-T90 has as depth Depth. Depth has as unit of measurement cm. Depth has as value float 9.4.

The Sony Cyber-shot DSC-T90 has as digital zoom factor the Digital Zoom Factor. The Digital Zoom Factor has as value float 12.1. [. . . ]

The Sony Cyber-shot DSC-T90 has as feature Video Recording, Microphone and the Automatic Picture Stabilizer.

The Sony Cyber-shot DSC-T90 has as self timer true. [. . . ]

In this ontology, many properties have composite values, expressed by using auxiliary individuals. In the example above, a property (`hasDepth`) connects the digital camera to an auxiliary individual `Depth` (similar to the anonymous node `_:n` of the property concatenation price example of page 691), which is then connected via two other properties (`hasValueFloat`

---

and `hasUnitOfMeasurement`) to the float value 9.4 and the unit of measurement (centimeters), respectively. We obtained the descriptions of the auxiliary individuals (e.g., `Depth`), which are different entries in the glossary of the SWAT verbalizer, and we copied them immediately after the corresponding sentences that introduce the auxiliary individuals. We also formatted each text as a list of sentences, as above, to improve readability.

We then generated texts for the 60 products using NaturalOWL(−), setting the maximum fact distance to 1. Descriptions of auxiliary individuals were also generated and copied immediately after the sentences introducing them. The texts were very similar to those of the SWAT verbalizer, and they were formatted in the same manner.

In this trial, we also wanted to consider a scenario where the set of individuals to be described changes frequently (e.g., the products sold by a reseller change, new products arrive etc.) along with changes in other connected individuals (e.g., new manufacturers may be added), but nothing else in the ontology changes, i.e., only the assertional knowledge changes. In this case, it may be impractical to update the domain-dependent generation resources whenever the population of individuals changes. Our hypothesis was that by considering a sample of individuals of the types to be described (printers, cameras, camcorders, in our case), it would be possible to construct domain-dependent generation resources (e.g., sections, the ordering of sections and properties, sentence plans, the NL names of classes) that would help NaturalOWL generate reasonably good descriptions of new (unseen) individuals (products), without updating the domain-dependent generation resources, using the tokenized OWL identifiers or `rdfs:label` strings of the new individuals as their NL names.

To simulate this scenario, we randomly split the 60 products in two non-overlapping sets, the *development set* and the *test set*, each consisting of 10 digital cameras, 10 camcorders, and 10 printers. Again, the second author constructed and refined the domain-dependent generation resources of NaturalOWL, this time by considering a version of the ontology that included the 30 development products, but not the 30 test products, and by viewing the generated texts of the 30 development products only. This took approximately six days (for two languages).[33] Hence, relatively light effort was again needed, compared to the time it typically takes to develop an ontology of this size, with terminology in two languages. Texts for the 30 products of the test set were then also generated by using NaturalOWL and the domain-dependent generation resources of the development set.

As in the previous trial, we defined only one user type, and we used interest scores only to block sentences stating the obvious. The maximum messages per sentence was again 3. We constructed domain-dependent generation resources for both English and Greek; the resources are summarized in Table 6. We created sentence plans only for the 42 properties of the ontology that were used in the development set (one sentence plan per property); the test set uses two additional properties, for which the default sentence plans of NaturalOWL (for English and Greek) were used. We also assigned the 42 properties to 6 sections, and ordered the sections and properties. We created NL names only when the automatically extracted ones were causing disfluencies in the development texts. Unlike the previous trial, the products to be described were not declared to be anonymous individuals, but the number of NL names that had to be provided was roughly the same as in the previous trial, since fewer automatically extracted names were causing disfluencies; in particular, all the products had reasonably good `rdfs:label` strings providing their English names.

---

33. Again, some of the domain-dependent generation resources were constructed by editing their OWL representations. As a test, the second author later reconstructed the domain-dependent generation resources from scratch using the fully functional Protégé plug-ing, this time in four days.

| Resources | English | Greek |
|---|---|---|
| Sections | 6 | |
| Property assignments to sections | 42 | |
| Interest score assignments | 12 | |
| Sentence plans | 42 | 42 |
| Lexicon entries | 19 | 19 |
| Natural language names | 36 | 36 |

Table 6: Domain-dependent generation resources for the **Consmer Electronics Ontology**.

An example description from the development set produced by NaturalOWL(+)follows. We formatted the sentences of each section as a separate paragraph, headed by the name of the section (e.g., "Other features:"); this was easy, because NaturalOWL can automatically mark up the sections in the texts. The maximum fact distance was again 1, but the sentence plans caused NaturalOWL to automatically retrieve additional message triples describing the auxiliary individuals at distance 1; hence, we did not have to retrieve this information manually, unlike the texts of the SWAT verbalizer and NaturalOWL(−).

**Type:** Sony Cyber-shot DSC-T90 is a digital camera.

**Main features:** It has a focal length range of 35.0 to 140.0 mm, a shutter lag of 2.0 to 0.0010 sec and an optical zoom factor of 4.0. It has a digital zoom factor of 12.1 and its display has a diagonal of 3.0 in.

**Other features:** It features an automatic picture stabilizer, a microphone, video recording and it has a self-timer.

**Energy and environment:** It uses batteries.

**Connectivity, compatibility, memory:** It supports USB 2.0 connections for data exchange and it has an internal memory of 11.0 GB.

**Dimensions and weight:** It is 5.7 cm high, 1.5 cm wide and 9.4 cm deep. It weighs 128.0 grm.

The 180 English texts that were generated by the three systems for the 30 development and 30 test products were shown to the same 10 students of the first trial. The students were now told that the texts would be used in on-line descriptions of products in the Web site of a retailer. Again, the OWL statements that the texts were generated from were not shown to the students, and the students did not know which system had generated each text. Each student was shown 18 randomly selected texts, 9 for products of the development set (3 texts per system) and 9 for products of the test set (again 3 texts per system). Each student was shown all of his/her texts in random order, regardless of the system that had generated them. The students were asked to score the texts as in the previous trial.

Table 7 shows the results for the English texts of the *development* set.[34] As in the previous trial, the domain-dependent generation resources clearly help NaturalOWL produce much more fluent sentences, and much better referring expressions and sentence orderings. The text structure scores of the SWAT verbalizer and NaturalOWL(−)are now much lower than in the previous trial, because there are now more message triples to express per individual and more topics, and the texts of these systems jump from one topic to another making the texts look very incoherent; for example, a sentence about the width of a camera may be separated from a sentence about its height by a sentence about shutter lag. This incoherence may have also contributed to the much lower clarity scores of these two systems, compared to the previous trial. The interest scores of these two systems are also much lower than in the previous trial; this may be due to the verbosity of their texts, caused by their frequent references to auxiliary individuals in the second trial, combined with the lack (or very little use) of sentence aggregation and pronoun generation. By contrast, the clarity and interest of NaturalOWL(+)were judged to be perfect; the poor clarity and interest of the other two systems may have contributed to these perfect scores though. Again, the SWAT verbalizer obtained slightly better scores than NaturalOWL *without* domain-dependent generation resources, except for clarity, but the differences are not statistically significant.

---

34. When a confidence interval is 0.00, this means that all the students gave the same score to all texts.

| Criteria | SWAT | NaturalOWL(−) | NaturalOWL(+) |
|---|---|---|---|
| Sentence fluency | $1.97 \pm 0.15$ | $1.93 \pm 0.27$ | $\mathbf{2.90 \pm 0.08}$ |
| Referring expressions | $1.10 \pm 0.06$ | $1.10 \pm 0.11$ | $\mathbf{2.87 \pm 0.08}$ |
| Text structure | $1.67 \pm 0.15$ | $1.33 \pm 0.19$ | $\mathbf{2.97 \pm 0.04}$ |
| Clarity | $1.97 \pm 0.15$ | $2.07 \pm 0.26$ | $\mathbf{3.00 \pm 0.00}$ |
| Interest | $1.77 \pm 0.14$ | $1.73 \pm 0.29$ | $\mathbf{3.00 \pm 0.00}$ |

Table 7: **English development** results for the **Consumer Electronics Ontology**.

| Criteria | SWAT | NaturalOWL(−) | NaturalOWL(+) |
|---|---|---|---|
| Sentence fluency | $2.03 \pm 0.15$ | $1.87 \pm 0.15$ | $\mathbf{2.87 \pm 0.08}$ |
| Referring expressions | $1.10 \pm 0.06$ | $1.10 \pm 0.06$ | $\mathbf{2.87 \pm 0.08}$ |
| Text structure | $1.57 \pm 0.13$ | $1.37 \pm 0.12$ | $\mathbf{2.93 \pm 0.05}$ |
| Clarity | $2.07 \pm 0.15$ | $1.93 \pm 0.15$ | $\mathbf{2.97 \pm 0.04}$ |
| Interest | $1.83 \pm 0.17$ | $1.60 \pm 0.14$ | $\mathbf{2.97 \pm 0.04}$ |

Table 8: **English test** results for the **Consumer Electronics Ontology**.

Table 8 shows the results for the English texts of the *test* set. The results of the SWAT verbalizer and NaturalOWL(−)are very similar to those of Table 7, as one would expect. Also, there was only a very marginal decrease in the scores of NaturalOWL(+), compared to the scores of the same system for the development set in Table 7. There is no statistically significant difference, however, between the corresponding cells of the two tables, for any of the three systems. These results support our hypothesis that by considering a sample of individuals of the types to be described one can construct domain-dependent generation resources that can be used to produce high-quality texts for new individuals of the same types, when the rest of the ontology remains unchanged. The fact that all the products (but not the other individuals) had `rdfs:label` strings providing their English names probably contributed to the high results of NaturalOWL(+)in the test set, but `rdfs:label` strings of this kind are common in OWL ontologies.

We then showed the 60 Greek texts that were generated by NaturalOWL(+)to the same 10 students, who were native Greek speakers; the SWAT verbalizer and NaturalOWL(−)cannot generate Greek texts from the Consumer Electronics ontology. Table 9 shows the results we obtained for the Greek texts of the development and test sets. There is no statistically significant difference from the corresponding results for English (cf. the last columns of Tables 7 and 8). There is also no statistically significant difference in the results for the Greek texts of the development and test sets (Table 9). We note, however, that it is common to use English names of electronics products in Greek texts, which made using the English `rdfs:label` names of the products in the Greek texts acceptable. In other domains, for example cultural heritage, it might be unacceptable to use English names of individuals; hence, one would have to provide Greek NL names for new individuals.

| Criteria | NaturalOWL(+), development data | NaturalOWL(+), test data |
|---|---|---|
| Sentence fluency | $2.87 \pm 0.12$ | $2.83 \pm 0.09$ |
| Referring expressions | $2.77 \pm 0.20$ | $2.80 \pm 0.11$ |
| Text structure | $3.00 \pm 0.00$ | $3.00 \pm 0.00$ |
| Clarity | $3.00 \pm 0.00$ | $2.93 \pm 0.05$ |
| Interest | $2.97 \pm 0.06$ | $3.00 \pm 0.00$ |

Table 9: **Greek** results for the **Consumer Electronics Ontology**.

| No. | System Configuration | Sentence Fluency | Ref. Expressions | Text Structure | Clarity | Interest |
|-----|---------------------|------------------|------------------|----------------|---------|----------|
| 1 | NaturalOWL(+) | **4.80 ± 0.12** | **5.00 ± 0.00** | **4.82 ± 0.15** | **4.78 ± 0.12** | **4.89 ± 0.09** |
| 2 | − interest scores | 4.53 ± 0.16 | 4.95 ± 0.06 | 4.78 ± 0.12 | 4.62 ± 0.17 | 4.20 ± 0.19 |
| 3 | − ref. expr. gen. | 3.93 ± 0.28 | *1.53 ± 0.22* | 4.80 ± 0.12 | 4.51 ± 0.24 | 4.07 ± 0.22 |
| 4 | − NL names | 3.71 ± 0.29 | 1.48 ± 0.21 | 4.71 ± 0.15 | 4.24 ± 0.25 | 3.98 ± 0.26 |
| 5 | − aggregation | 3.64 ± 0.33 | 1.33 ± 0.19 | 4.67 ± 0.16 | 4.24 ± 0.25 | 3.93 ± 0.26 |
| 6 | − sentence plans | *2.07 ± 0.37* | 1.33 ± 0.19 | 4.60 ± 0.18 | *2.49 ± 0.36* | *2.38 ± 0.35* |
| 7 | − sections, ordering | 1.89 ± 0.36 | 1.33 ± 0.19 | *1.53 ± 0.24* | 2.33 ± 0.33 | 1.89 ± 0.28 |

**Table 10:** **Ablation** English test results for the **Consumer Electronics Ontology**. Each configuration removes one component or resource from the previous configuration.

## 4.3 Ablation Trials with the Consumer Electronics Ontology

In the last trial, we studied how the quality of the generated texts is affected when various components and domain-dependent generation resources of NaturalOWL are gradually removed. We used the Consumer Electronics Ontology, with the domain-dependent generation resources that we had constructed for the 30 development products of the previous trial. We also used 45 new test products (15 digital cameras, 15 camcorders, and 15 printers, from the same publicly available dataset), other than the 30 development and the 30 test products of the previous trial.

We generated English texts for the 45 new test products, using the 7 configurations of NaturalOWL of Table 10. The resulting $45 \times 7 = 315$ texts were shown to 7 students, who had the same background as in the previous trials. Each student was shown the 7 texts of 6 or 7 test products (42 or 49 texts per student). For each product, the 7 texts were shown side by side in random order, and the students were instructed to take into account the differences of the 7 texts. The students did not know which system had generated which text. The same criteria (statements $S_1$–$S_5$ of Section 4.1) were used again, but a scale from 1 to 5 was used this time (1: strong disagreement, 2: disagreement, 3: ambivalent, 4: agreement, 5: strong agreement), to make it easier to distinguish between the 7 configurations.

The first configuration (NaturalOWL(+)) is NaturalOWL with all of its components enabled, using all the available domain-dependent generation resources. As in the previous trial (see Table 8), the texts of this configuration were judged to be near-perfect by all the criteria. The second configuration was the same, but without the interest score assignments. The results of the second configuration were very close to the results of the first one, since interest score assignments were used only to avoid generating sentences stating the obvious (e.g., "Sony Cyber-shot DSC-T90 is manufactured by Sony"). The biggest decrease was in the interest criterion, as one would expect, but the scores for sentence fluency and clarity were also affected, presumably because the sentences that state the obvious sound unnatural and seem to introduce noise. There were very small differences in the scores for referring expressions and text structure, which seem to suggest that when the overall quality of the texts decreases, the judges are biased towards assigning lower scores in all of the criteria.[35]

The third configuration was the same as the second one, but the component that generates pronouns and demonstrative noun phrases was disabled, causing NaturalOWL to always use the NL names of the individuals and classes, or names extracted from the ontology. There was a big decrease in the score for referring expresions, showing that despite their simplicity, the referring expression generation methods of NaturalOWL have a noticeable effect; we mark big decreases in italics in Table 10. The scores for sentence fluency, interest, and clarity were also affected, presumably because repeating the names of the individuals and classes made the sentences look less natural, boring, and more difficult to follow. There was almost no difference (a very small positive one) in the text structure score.

In the fourth configuration, the NL names of the individuals and classes were also removed, forcing NaturalOWL to always use automatically extracted names. There was a further decrease in the score for referring expressions, but the decrease was small, because the referring expressions were already poor in the third configuration. Note, also, that the NL names are necessary for NaturalOWL to produce pronouns and demonstrative noun phrases; hence, the higher referring expression score of the third configuration would not have been possible without the NL names. The sentence fluency and clarity scores were also affected in the fourth configuration, presumably because the automatically extracted names made the texts more difficult to read and understand. There were also small decreases in the scores for interest and even text structure, suggesting again that when the overall quality of the texts decreases, the judges are biased towards lower scores in all of the criteria.

In the fifth configuration, aggregation was turned off, causing NaturalOWL to produce a separate sentence for each message triple. With sentences sharing the same subject no longer being aggregated, more referring expressions for subjects had to be generated. Since the component that generates pronouns and demonstrative noun phrases had been switched off and the NL names had been removed, more repetitions of automatically extracted names had to be used, which is why the score for referring expressions decreased further. Sentence fluency was also affected, since some obvious aggregations were no longer being made, which made the sentences look less natural. There was also a small decrease in the score for the perceived text structure and interest, but no difference in the score for clarity. Overall, the contribution of aggregation to the perceived quality of the texts seems to be rather small.

In the sixth configuration, all the sentence plans were removed, forcing NaturalOWL to use the default sentence plan and tokenized property identifiers. There was a sharp decrease in sentence fluency and clarity, as one would expect, but also in the perceived interest of the texts. There was also a small decrease in the perceived text structure, and no difference in the score for referring expressions. Overall, these results indicate that sentence plans are a very important part of the domain-dependent generation resources.

In the seventh configuration, the sections, assignments of properties to sections, and the ordering of sections and properties were removed, causing NaturalOWL to produce random orderings of the message triples. There was a very sharp decrease in the score for text structure. The scores for the perceived interest, clarity, but also sentence fluency were also affected, again suggesting that when the overall quality of the texts decreases, the judges are biased towards lower scores in all of the criteria.

---

35. In all of the criteria, all the differences from one configuration to the next one are statistically significant, with the only exceptions being the differences in clarity between configurations 4 and 5, and the differences in the scores for referring expressions between configurations 5–6 and 6–7. Again, when the 95% confidence intervals overlapped, we performed paired two-tailed $t$-tests ($\alpha = 0.05$).

We conclude that the sections and ordering information of the domain-dependent generation resources are, along with the sentece plans, particularly important. We note, however, that the best scores were obtained by enabling all the components and using all the available domain-dependent generation resources.

## 5. Conclusions and Future Work

We provided a detailed description of NaturalOWL, an open-source NLG system that produces English and Greek texts describing individuals or classes of OWL ontologies. Unlike simpler verbalizers, which typically express a single axiom at a time in controlled, often not entirely fluent English primarily for the benefit of domain experts, NaturalOWL aims to generate fluent and coherent multi-sentence texts for end-users in more than one languages.

We discussed the processing stages of NaturalOWL, the optional domain-dependent generation resources of each stage, as well as particular NLG issues that arise when generating from OWL ontologies. We also presented trials we performed to measure the effort required to construct the domain-dependent generation resources and the extent to which they improve the resulting texts, also comparing against a simpler OWL verbalizer that requires no domain-dependent generation resources and employs NLG methods to a lesser extent. The trials showed that the domain-dependent generation resources help NaturalOWL produce significantly better texts, and that the resources can be constructed with relatively light effort, compared to the effort that is typically needed to develop an OWL ontology.

Future work could compare the effort needed to construct the domain-dependent generation resources against the effort needed to manually edit the lower quality texts produced without domain-dependent generation resources. Our experience is that manually editing texts generated by a verbalizer (or NaturalOWL(−)) is very tedious when there is a large number of individuals (e.g., products) of a few types to be described, because the editor has to repeat the same (or very similar) fixes. There may be, however, particular applications where post-editing the texts of a simpler verbalizer may be preferable.

We also aim to replace in future work the pipeline architecture of NaturalOWL by a global optimization architecture that will consider all the NLG processing stages in parallel, to avoid greedy stage-specific decisions (Marciniak & Strube, 2005; Lampouras & Androutsopoulos, 2013a, 2013b). Finally, we hope to test NaturalOWL with biomedical ontologies, such as the Gene Ontology and SNOMED.[36]

---

36. See `http://www.geneontology.org/` and `http://www.ihtsdo.org/snomed-ct/`.